\documentclass[journal]{IEEEtran}

\usepackage{times}
\usepackage[marginal]{footmisc}
\usepackage{graphicx}
\usepackage{epstopdf}
\usepackage{epsfig}
\usepackage{amsmath,bm}
\usepackage{amssymb}
\usepackage{algorithm}
\usepackage{algorithmic}
\usepackage{array}
\usepackage{color}
\usepackage{booktabs}
\usepackage{multirow}
\usepackage{verbatim}
\usepackage[colorlinks, urlcolor=black, linkcolor=black, anchorcolor=black, citecolor=black]{hyperref}
\usepackage{cite}

\ifCLASSINFOpdf
\else
\fi
\begin{document}
\title{Two-stream Collaborative Learning with Spatial-Temporal Attention for Video Classification}

\author{Yuxin~Peng,
        Yunzhen~Zhao,
        and Junchao~Zhang
\thanks{This work was supported by National Natural Science Foundation of China under Grants 61771025 and 61532005.}
\thanks{The authors are with the Institute of
Computer Science and Technology, Peking University, Beijing 100871,
China. Corresponding author: Yuxin Peng. (E-mail: pengyuxin@pku.edu.cn)}}%

\maketitle

\begin{abstract}

Video classification is highly important with wide applications, such as video search and intelligent surveillance. Video naturally consists of static and motion information, which can be represented by frame and optical flow. Recently, researchers generally adopt the deep networks to capture the static and motion information \textbf{\emph{separately}}, which mainly has two limitations:
(1) Ignoring the coexistence relationship between spatial and temporal attention, while they should be jointly modelled as the spatial and temporal evolutions of video, thus discriminative video features can be extracted.
(2) Ignoring the strong complementarity between static and motion information coexisted in video, while they should be collaboratively learned to boost each other.
For addressing the above two limitations, this paper proposes the approach of two-stream collaborative learning with spatial-temporal attention (TCLSTA), which consists of two models:
(1) \emph{Spatial-temporal attention model}: The spatial-level attention emphasizes the salient regions in frame, and the temporal-level attention exploits the discriminative frames in video. They are jointly learned and mutually boosted to learn the discriminative static and motion features for better classification performance.
(2) \emph{Static-motion collaborative model}: It not only achieves mutual guidance on static and motion information to boost the feature learning, but also adaptively learns the fusion weights of static and motion streams, so as to exploit the strong complementarity between static and motion information to promote video classification.
Experiments on 4 widely-used datasets show that our TCLSTA approach achieves the best performance compared with more than 10 state-of-the-art methods.

\end{abstract}

\IEEEpeerreviewmaketitle

\section{Introduction}

Video classification is a highly important task, and has wide applications such as video search, intelligent surveillance, human-computer interaction and elderly care. A recent statistical study shows that video traffic will be $82$ percent of all consumer Internet traffic by 2021 \footnote{https://www.cisco.com/c/en/us/solutions/collateral/service-provider/visual-networking-index-vni/complete-white-paper-c11-481360.html}. Under this situation, video classification is urgently required. It has drawn extensive attention over past several decades, and many works have been proposed for effective approaches and representative benchmarks \cite{15.simonyan2014two, 9.wu2015modeling, 87.zhao2017pooling}.

The main challenges of video classification come from three aspects:
(1) A large portion of video categories have unconstrained content involving various objects. For example, many objects (e.g. flowers, cakes, knifes) appear in the scene of ``birthday party", which make the video content of this category very complex and hard to be recognized.
(2) Videos belonging to the same category may have greatly different context information. For example, ``birthday party" can be held in a dinning hall with tables, or in a backyard with green grass.
(3) Videos belonging to different categories may have similar content. For example, ``soccer juggling" and ``soccer penalty" share the same content of soccer, athletes, and green grass.
These challenges lead to great difficulty in video classification.
Since video naturally consists of static and motion information, traditional video classification methods directly use hand-crafted features to model these two types of information, such as histograms of oriented gradients (HOG) \cite{18.dalal2005histograms} for static information and histograms of optical flow (HOF)\cite{115.dalal2006human} for motion information.
\begin{figure*}[!t]
  \centering
  \includegraphics[width=1.0\textwidth]{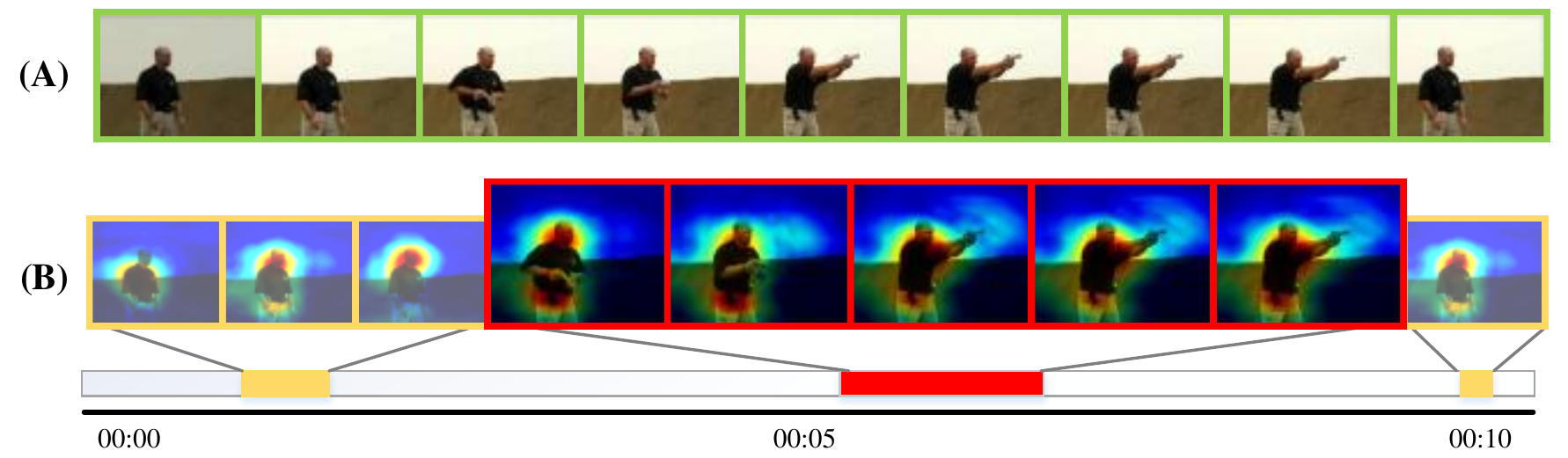}
  \caption{An example of spatial-temporal attention for a video sequence of ``Shoot gun". (A) shows the original video sequence. In (B), spatial-level attention is shown by heatmap for each frame, and frames with the red rectangles and bigger size are with higher temporal-level attention, which focus on the action sequence of shooting.}
  \label{fig:attention}
\end{figure*}

Recently, due to the strong power in feature learning, deep learning has been widely applied to video classification for modelling static and motion information.
Deep networks automatically learn a hierarchy of features from large scale raw data using a general-purpose learning procedure, which leads to discriminative features with high abstraction and invariance \cite{lecun2015deep,bengio2013representation}. While hand-crafted features don't have above advantages as deep networks, thus have limited discriminative capacity \cite{57.wang2015action,xie2017hybrid,23.yue2015beyond}.
Simonyan et al. \cite{15.simonyan2014two} employ two convolutional neural networks (CNNs) to model static and motion information separately, which take frame and optical flow as input, and achieve better performance than the traditional video classification methods.
Inspired by \cite{15.simonyan2014two}, some methods \cite{9.wu2015modeling,23.yue2015beyond} are further proposed to apply two networks to model static and motion information.
Despite achieving promising performance, these works mainly \textbf{\emph{separately}} model static and motion information. However, for the coexistence relationship between static and motion information, they provide complementary clues for the same video category, thus should be collaboratively learned to promote the feature learning. This is the \textbf{first limitation}.

In addition, these works ignore spatial-temporal attention, which is very important for video classification. On \emph{\textbf{spatial-level attention}}, different regions in frame have different degrees of saliency, where salient regions should be assigned more attention.
On \emph{\textbf{temporal-level attention}}, different frames in video sequence contribute to video classification differently, where discriminative frames should be paid more attention to. 
Fig. \ref{fig:attention} shows an example of spatial-temporal attention for a video sequence of ``Shoot gun". Fig. \ref{fig:attention}.(A) shows the original video sequence, and Fig. \ref{fig:attention}.(B) shows spatial-temporal attention. In Fig. \ref{fig:attention}.(B), spatial-level attention is shown by heatmaps, in which regions including the man and gun are with more spatial-level attention, while frames with red rectangles and bigger sizes are with more temporal-level attention, which are more discriminative for semantic representation of ``Shoot gun".
Recent years, some methods are proposed to learn spatial-level attention by CNNs \cite{62.jaderberg2015spatial} or RNNs (recurrent neural networks) \cite{60.mnih2014recurrent}, and utilize the conditional entropy of visual words \cite{50.zhao2008information} or AdaBoost \cite{51.liu2013boosted} to measure temporal-level attention.
However, existing works learn spatial-level attention \textbf{\emph{or}} temporal-level attention \textbf{\emph{separately}}, and ignore their coexistence relationship, which cannot fully exploit the discriminations of spatial-level and temporal-level attention.
This is the \textbf{second limitation}.

\begin{figure}[tbp]
  \centering
  \includegraphics[width=0.48\textwidth]{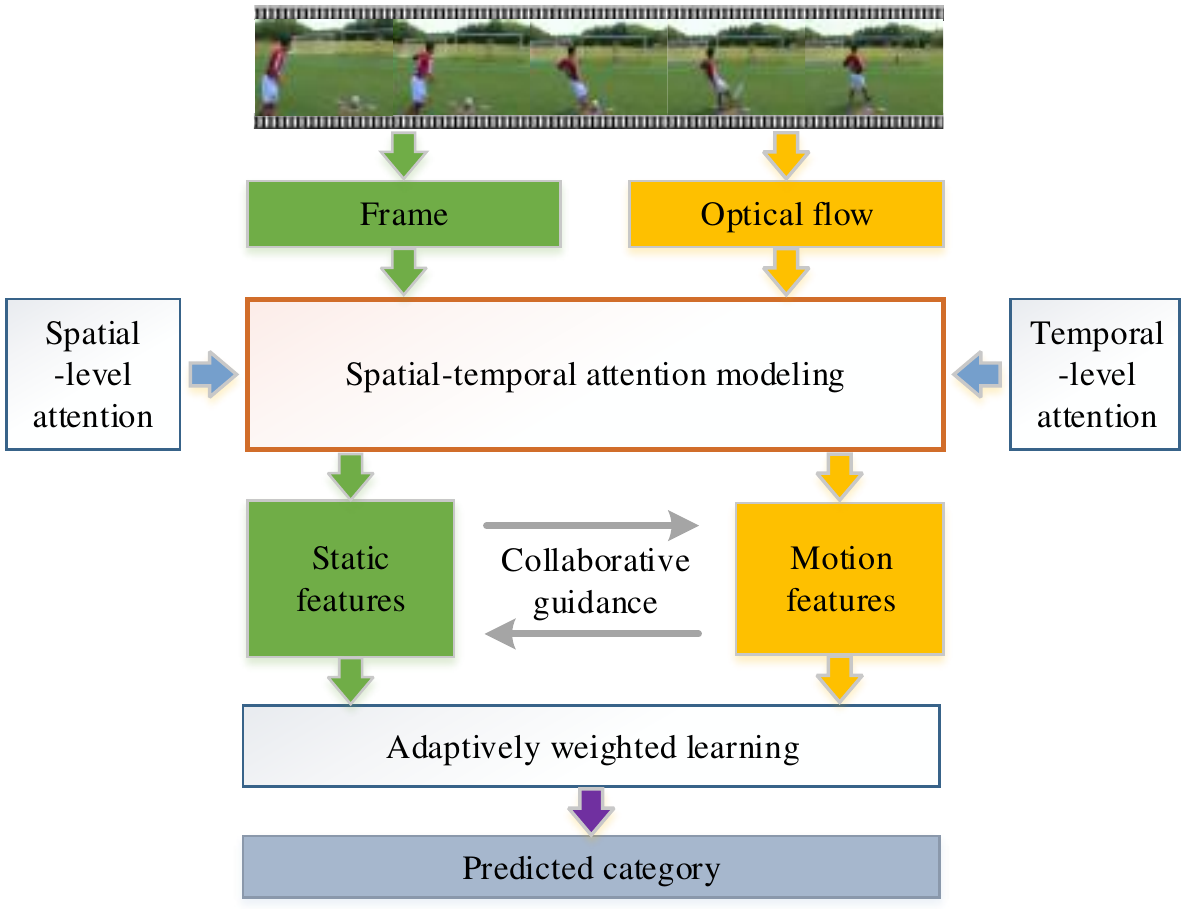}
  \caption{The design idea of our proposed TCLSTA approach.}
  \label{fig:design_idea}
  \label{fig:lstm}
\end{figure}

For addressing these two limitations, this paper proposes the approach of two-stream collaborative learning with spatial-temporal attention (TCLSTA) for video classification. It first models the spatial-temporal attention to emphasize salient regions in frame and exploit discriminative frames, then performs collaborative guidance on static and motion information and adaptively learns fusion weights of static and motion streams to exploit strong complementarity between them for improving the classification accuracy, as shown in Fig. \ref{fig:design_idea}.
The main contributions of our TCLSTA approach are summarized as follows:
\begin{itemize}
\renewcommand{\labelitemi}{$\vcenter{\hbox{\scriptsize$\bullet$}}$}
\item \textbf{\emph{Static-motion collaborative model}}. Existing works model the static and motion information separately \cite{15.simonyan2014two,9.wu2015modeling,23.yue2015beyond}, which ignore the strong complementarity between them. However, video naturally contains static and motion information, which are two complementary aspects to represent the same semantic category, and can provide important cues for each other to guide the feature learning. For addressing this problem, we propose a \textbf{\emph{static-motion collaborative model}} to jointly exploit the discriminative static and motion information using an alternate training scheme. On the one hand, it allows mutual guidance between static and motion information to promote feature learning, which exploits strong complementarity for learning discriminative static and motion features. On the other hand, it adaptively learns the fusion weights of static and motion streams, which distinguishes different roles of static and motion information for each category to improve classification accuracy.
\item \textbf{\emph{Spatial-temporal attention model}}. Existing works learn the spatial-level attention or temporal-level attention separately \cite{60.mnih2014recurrent,50.zhao2008information,51.liu2013boosted}, which ignore the coexistence relationship of them. However, the spatial locations of salient regions in video sequence vary over time, thus temporal-level attention can guide spatial-level attention learning to focus on the regions in discriminative frames, while spatial-level attention can guide temporal-level attention learning to emphasize the frames with discriminative objects, so they can greatly boost each other and should be jointly modelled. For addressing this problem, we propose a \textbf{\emph{spatial-temporal attention model}} to jointly capture the video evolutions both in spatial and temporal domains. It achieves mutual boosting on emphasizing salient regions and highlighting discriminative frames, which can learn discriminative features that combine spatial-level and temporal-level attention.
\end{itemize}

Extensive experiments are conducted on 4 widely-used datasets to verify the performance of our approach, which show that our TCLSTA approach achieves the best performance compared with more than 10 state-of-the-art methods.
The rest of this paper is organized as follows. Section II introduces the related works, Section III presents the proposed approach in detail, and then in Section IV experimental results and analyses are shown, followed by conclusions and future works in Section V.

\section{Related Works}

Video classification has been an attractive research topic, and achieved great progress in recent years \cite{9.wu2015modeling,2.wang2013action,15.simonyan2014two}. In the following, we first review the existing works on video classification in two aspects: feature representation and visual attention, then we briefly review related works on alternate training scheme, which is used in our TCLSTA approach.

\subsection{Feature representation}
\subsubsection{Methods based on hand-crafted features}
We discuss the hand-crafted features for video representation from three aspects: low-level features, mid-level features and high-level features. Low-level features focus on describing visual patterns of local informative regions or 3D volumes.
Among methods of \emph{low-level} features,
Laptev \cite{16.laptev2005space} extends Harris detector to Harris3D detector to detect spatial-temporal interest points in video. The histograms of oriented gradients (HOG) \cite{18.dalal2005histograms} are used to capture static appearance information, while histograms of optical flow (HOF) \cite{115.dalal2006human} and motion boundary histograms (MBH) \cite{115.dalal2006human} are proposed to capture local motion information. Wang et al. \cite{2.wang2013action} further make use of point trajectories and construct effective video representation that combines HOG, HOF, MBH and trajectories. Xian et al. \cite{90.xian2017evaluation} conduct comprehensive experiments to evaluate the performance of low-level features. 
Besides low-level features, researchers also resort to mid-level and high-level features to construct video representations, which usually select discriminative feature units or utilize high-level semantic concepts.
Among methods of \emph{mid-level} features,
Wang et al. \cite{wang2013mining} introduce motion atom and motion phrase to represent complex actions in video, where motion atom is in short temporal scale and motion phase is in long temporal scale. Both of them can be seen as mid-level ``parts'' of complex actions.
Liu et al. \cite{17.liu2011recognizing} propose to represent videos by a set of intermediate concepts, where the intermediate concepts are either manually specified or learnt from the training data.
Wang et al. \cite{wang2013motionlets} construct mid-level video representations by mining spatial-temporal part with coherent appearance and motion features, which can provide a tradeoff between repeatability and discriminative ability.
With considering both individual and interactive actions in video, Fradi et al. \cite{fradi2017crowd} propose to extract a rich set of mid-level visual descriptors from compact local representations and utilize them to encode semantic information.
Among methods of \emph{high-level} features,
Izadinia et al. \cite{izadinia2012recognizing} first learn several concepts from manually labeled data, and then construct video representations according to the responses of concept detectors.
Sun et al. \cite{sun2013active} use the Hidden Markov Model to capture the temporal transitions between concepts in video, and encode the video into a fixed length vector by Fisher Vector.
Zhang et al. \cite{89.zhang2017exploring} propose to extract coherent
motion features using a structured trajectories learning method, and present a high-level crowd motion behavior representation to describe the high-level semantic information about crowd behaviors.

\subsubsection{Methods based on deep features}

Nowadays, deep learning has shown its strong power in feature learning. Deep networks have the ability of learning more discriminative and robust features \cite{xie2017hybrid,23.yue2015beyond}, and achieves great progress in video classification \cite{15.simonyan2014two,wu2016multi,11.ji20133d,10.tran2014learning,56.sun2015human,wang2016temporal,106.donahue2015long,103.feichtenhofer2016spatiotemporal,9.wu2015modeling,yan2014modeling}.

Following \cite{herath2017going}, according to the difference of network design, we summarize methods based on deep features into 4 groups, namely methods based on spatial-temporal networks, multiple stream networks, deep generative networks and temporal coherency networks. A summarization of these 4 groups is presented as Table \ref{tab:deepsummary}.
The methods based on \emph{spatial-temporal networks} \cite{11.ji20133d,10.tran2014learning,23.yue2015beyond,56.sun2015human,wang2016temporal,106.donahue2015long} adopt deep networks to extract features from both spatial and temporal dimensions. Among them, Ji et al. \cite{11.ji20133d} extend 2D convolutional kernels to 3D kernels, which enable 3D CNNs to extract features from both spatial and temporal dimensions, thus capture spatial and temporal information in video. Ng et al. \cite{23.yue2015beyond} utilize max pooling strategy and long short-term memory network (LSTM) to combine image information across a video over long time periods, which aggregates spatial and temporal information to achieve better video classification accuracy. Wang et al. \cite{wang2016temporal} propose temporal segmentation network for video classification, which is a video-level framework and can model long-term temporal structure. Due to its segmental architecture with sparse sampling, this work promotes the video classification accuracy while maintaining a reasonable computational cost.
The methods based on \emph{multiple stream networks} \cite{15.simonyan2014two,wu2016multi,53.feichtenhofer2016convolutional,103.feichtenhofer2016spatiotemporal,9.wu2015modeling} adopt deep networks that consist of more than one stream to accept inputs in multiple modalities, such as frame, optical flow and audio, which are fused to get the final prediction. Simonyan et al. \cite{15.simonyan2014two} propose a two-stream CNN architecture for video classification, which consists of two CNNs that take frame and optical flow as input respectively. Besides frame and optical flow, Wu et al. \cite{9.wu2015modeling} further exploit audio signal to improve the video classification performance.
The methods based on \emph{deep generative networks} \cite{yan2014modeling, 55.srivastava2015unsupervised} adopt generative models for feature learning from temporal sequence in an unsupervised fashion. Yan et al. \cite{yan2014modeling} introduce a deep auto-encoder, denoted as Dynencoder, to capture video dynamics, which is learned in an unsupervised fashion with two stages, a layer-wise pre-training stage and an end-to-end finetuning stage. Srivastava et al. \cite{55.srivastava2015unsupervised} introduce an LSTM-based auto-encoder including an encoder LSTM and a decoder LSTM to discover long-term cues from video sequence.
The methods based on \emph{temporal coherency networks} \cite{goroshin2015unsupervised, wang2016actions} take the assumption that video frames are correlated both semantically and dynamically, where the temporal coherency means that the frames are in the correct temporal order.
For example, Wang et al. \cite{wang2016actions} split video frames about an action or event into precondition set and effect set, and then the video is identified by learning transformations from precondition set to effect set.

These works mainly \textbf{\emph{separately}} learn the feature representations of static and motion information. However, according to \cite{15.simonyan2014two}, static information from individual frames and motion information across frames naturally coexist in video, which are two essentially complementary aspects to represent a semantic category, thus should be collaboratively learned to boost each other. For example, for the video category ``Basketball Dunk'', the static information such as basketball, basketry and player can help to distinguish it from other categories unrelated to basketball sport, and the motion information such as player's action can further distinguish it from the categories related to basketball sport but with different actions, such as shooting basketball. For exploiting the strong complementarity between static and motion information, we propose a static-motion collaborative model, which jointly learns the discriminative static and motion features, thus can mutually boost the representation and optimize the fusion weights of frame and optical flow.

\begin{table}[htb]
\newcommand{\tabincell}[2]{\begin{tabular}{@{}#1@{}}#2\end{tabular}}
\centering
\caption{Summarization of methods based on deep features.}
\label{tab:deepsummary}
\begin{tabular}{|c|c|p{4.1cm}|}
\hline
Group & Methods & \hspace{1.1cm}Summarization \\
\hline
\multirow{3}{*}{\tabincell{c}{Spatial-temporal\\networks}}&\multirow{3}{*}{\tabincell{c}{\cite{23.yue2015beyond} \cite{11.ji20133d}\\\cite{10.tran2014learning} \cite{56.sun2015human}\\\cite{wang2016temporal} \cite{106.donahue2015long}}}& Adopt deep networks to extract features from both spatial and temporal dimensions. \\
\hline
\multirow{5}{*}{\tabincell{c}{Multiple stream\\networks}} &\multirow{5}{*}{\tabincell{c}{\cite{15.simonyan2014two}\cite{wu2016multi}\\\cite{9.wu2015modeling}\cite{103.feichtenhofer2016spatiotemporal}\\\cite{53.feichtenhofer2016convolutional}}}& Adopt deep networks that consist of more than one stream to accept inputs in multiple modalities, such as frame, optical flow and audio, which are fused to get the final prediction. \\
\hline
\multirow{3}{*}{\tabincell{c}{Deep generative\\networks}} &\multirow{3}{*}{\tabincell{c}{\cite{yan2014modeling}\cite{55.srivastava2015unsupervised}}}& Adopt generative models for feature learning from temporal sequence in an unsupervised fashion. \\
\hline
\multirow{5}{*}{\tabincell{c}{Temporal coherency\\networks}} &\multirow{5}{*}{\tabincell{c}{\cite{goroshin2015unsupervised}\cite{wang2016actions}}}& Train networks by exploiting temporal orders to learn visual representations, which takes the assumption that video frames are correlated both semantically and dynamically. \\
\hline
\end{tabular}
\end{table}

\subsection{Methods based on visual attention}
\subsubsection{Spatial-level attention}
Regions in frame have different contributions to video classification, which should be assigned different degrees of attention.
Recently, some researchers introduce spatial-level attention models into video classification.
Mnih et al. \cite{60.mnih2014recurrent} present a RNN based method to extract visual attention from videos, which adaptively selects a sequence of regions and only processes the selected regions with high resolution.
Karpathy et al. \cite{7.karpathy2014large} design a multi-resolution CNN with fixing the attention at the center of frame. However, the salient regions are not always located at the center of frame.
Jaderberg et al. \cite{62.jaderberg2015spatial} add a soft attention mechanism between layers of CNNs. Instead of weighting locations by a softmax layer, they apply affine transformations to multiple layers.
\subsubsection{Temporal-level attention}
Temporal-level attention focuses on highlighting discriminative frames that contribute more to semantic representation of video.
Zhao et al. \cite{50.zhao2008information} present an approach to find the discriminative frames in video sequence, and represent them with the distribution of local motion features. But they use the fixed threshold to select the key-frames, which is not robust enough. 
Liu et al. \cite{51.liu2013boosted} select discriminative frames based on boosted frame selection. They use a supervised pyramidal motion feature which combines optical flow with a biologically inspired feature to detect interest points.
Barrett et al. \cite{92.barrett2016action} propose to automatically identify the most discriminative temporal subsequences in video, which is used for automatically learning the changing appearance and motion patterns of actions.

\begin{figure*}[!t]
  \centering
  \includegraphics[width=1.02\textwidth]{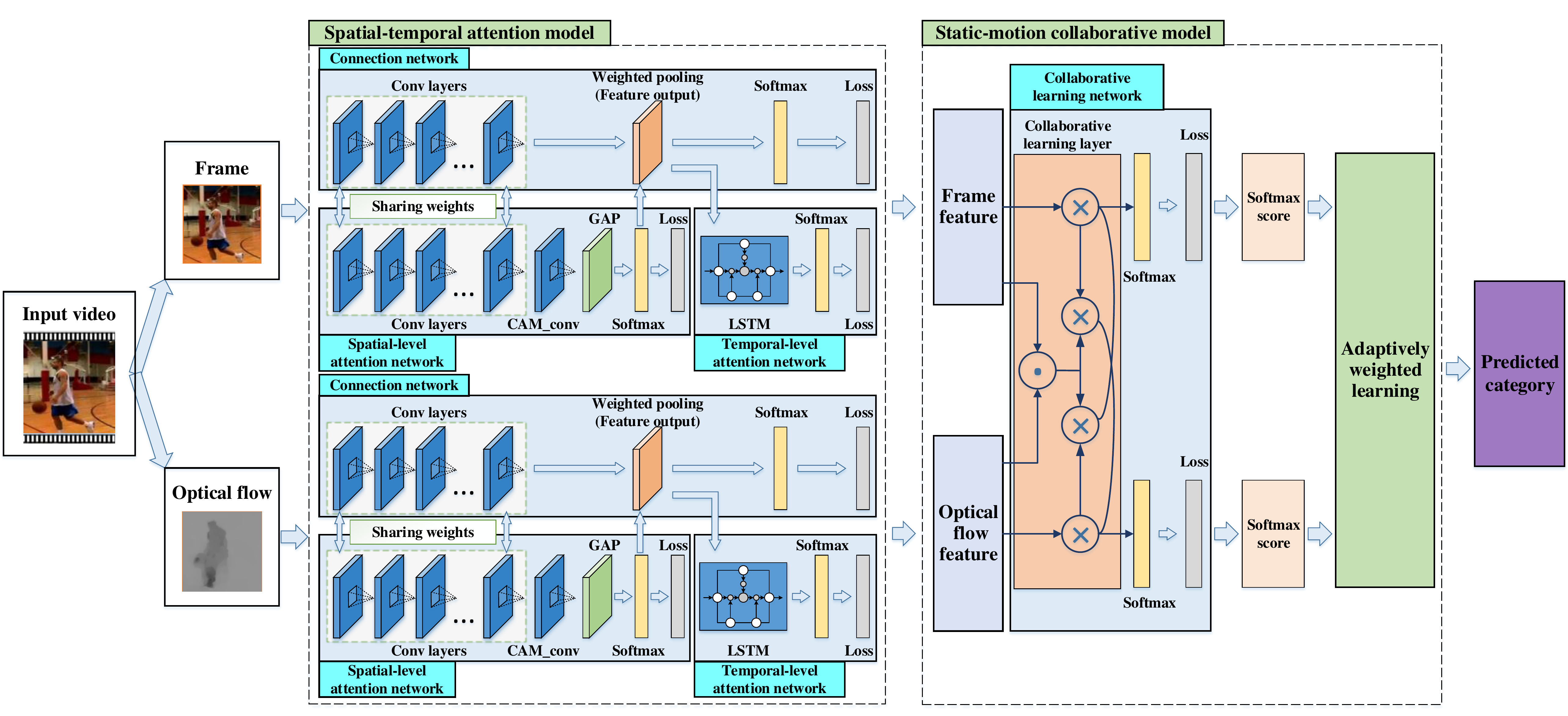}
  \caption{Our proposed TCLSTA framework consists of two components: \textbf{spatial-temporal attention model} is proposed to extract discriminative features from frame (static) and optical flow (motion) by jointly learning spatial-temporal attention in video, and \textbf{static-motion collaborative model} is proposed to exploit the strong complementarity between static and motion information for improving video classification performance. Each loss is cross-entropy loss.}
  \label{fig:framework}
\end{figure*}

However, above works learn spatial-level attention \textbf{\emph{or}} temporal-level attention \textbf{\emph{separately}}, while ignore their coexistence relationship. In this paper, we propose a spatial-temporal attention model, which jointly learns the spatial-level attention and temporal-level attention, and boosts both of them to improve the accuracy of video classification.

\subsection{Alternate training scheme}
The alternate training scheme is often used to optimize multi-task networks \cite{lu2016hierarchical, su2017multi, tang2017collaborative}, which is proposed to alternatively optimize objective functions of multiple tasks in a manner of optimizing one of them while fixing the others. Lu et al. \cite{lu2016hierarchical} apply the alternate training scheme to achieve an alternating co-attention mechanism for addressing visual question answer (VQA) problem, which improves the VQA performance by jointly reasoning about visual attention and question attention. Su et al. \cite{su2017multi} propose a multi-task learning method to address the problem of person re-identification on multi-cameras, and they adopt an alternating optimization strategy to solve the objective functions. Tang et al. \cite{tang2017collaborative} investigate a multi-task collaborative learning approach with an alternate manner, which is applied for joint learning of speech and speaker recognition tasks.
Inspired by these works, we apply the alternate training scheme for effectively driving the training of collaborative learning network in our approach, in order to allow mutual guidance on the optimization of static and motion features. 

\section{Our TCLSTA approach}
In this section, we will introduce the proposed TCLSTA approach in detail, which consists of two models: spatial-temporal attention model and static-motion collaborative model, as shown in Fig. \ref{fig:framework}. 
Input videos are first decomposed into frame and optical flow for representing static and motion information. Then spatial-temporal attention model extracts discriminative static and motion features from frame and optical flow by jointly modelling spatial-temporal attention. Finally, the static-motion collaborative model is proposed to optimize the static and motion features and adaptively learn the fusion weight for each video category.

\subsection{Spatial-temporal attention model}
This subsection introduces our proposed spatial-temporal attention model. As shown in Fig. \ref{fig:framework}, the proposed model consists of two streams, which take frame and optical flow as inputs respectively. Each stream is composed of three parts: connection network, spatial-level attention network, and temporal-level attention network. The spatial-level and temporal-level attention networks emphasize salient regions in frame and discriminative frames in video respectively, and the connection network learns discriminative features with spatial-temporal attention. The details of above three networks are presented as follows:

\subsubsection{Connection network}
The connection network is constructed based on ResNet-50 \cite{95.he2016deep} and we make the following changes to connect spatial-level and temporal-level attention networks: (1) We use a weighted pooling layer to replace the last pooling layer in original ResNet-50, which is connected to the softmax layer in spatial-level attention network, so that the spatial-level attention can be exploited to guide the feature learning. (2) The feature output layer (weighted pooling layer) is connected to the temporal-level attention network, and serves as its input, so that the connection network and temporal-level attention network can influence each other through the joint training procedure.
\subsubsection{Spatial-level attention network}
A video frame can be decomposed into salient and non-salient regions. ``Salient regions'' contain information of discriminative objects and distinct motion patterns\cite{li2016visual, wang2015consistent, mauthner2015encoding}. These regions provide indications for visual foregrounds and important information associated to pre-defined semantic categories. We denote surrounding background areas as ``non-salient regions'', which are less relevant to the semantic categories. For example, for a video of ``Horse Riding'' category in UCF101 dataset, the salient regions contain the horse and the rider, which indicate the discriminative objects and distinct motion patterns for ``Horse Riding'' category. While the background areas are non-salient regions that are less relevant to the ``Horse Riding'' category.

We first use a spatial-level attention network to extract the salient regions by adopting category activation mapping (CAM) \cite{52.zhou2016learning}, and then propose a weighted pooling method to guide the learning of connection network, thus the features with spatial attention information can be extracted from connection network.

As shown in Fig. \ref{fig:framework}, spatial-level attention network is constructed based on connection network, and they share weights on the convolutional layers.
Following \cite{52.zhou2016learning}, another convolutional layer (\emph{CAM\_conv}), global average pooling (GAP) layer and softmax layer are deployed after the shared convolutional layers.
The weights of the softmax layer are propagated back to the convolution layers for identifying the importance of different regions.

Formally, for a given frame, we denote the activation of unit $k$ in \emph{CAM\_conv} layer at location $(x, y)$ as $a_k(x, y)$.
Then for a certain category $c$, we denote the corresponding weight of unit $k$ and the corresponding input of softmax layer as $w_k^c$ and $s_c$ respectively. We can obtain that
\begin{equation}
\label{equ:sc}
  s_c = \sum_{x, y} \sum_k w_k^c a_k(x, y) = \sum_{x, y} m_c(x, y)
\end{equation}
\begin{equation}
  m_c (x, y) = \sum_k w_k^c a_k(x, y)
\end{equation}
where $s_c$ can indicate the overall \emph{importance} of convolutional activations for category $c$, thus $m_c(x, y)$ can directly indicate the \emph{importance} of the activation at spatial location $(x, y)$ 
for category $c$. So we defined $m_c(x,y)$ as the spatial-level attention.
Similar to \cite{yang2016discovering}, we normalize $m_c(x, y)$ to the same scale as
\begin{equation}
\label{equ:spatial_attention}
  \widetilde{m}_c{(x, y)} = \frac {g \cdot exp(m_c{(x, y)})} {\sum_{x, y} exp(m_c{(x, y))}}
\end{equation}
where $g$ stands for the number of corresponding spatial grids in a frame.

The obtained spatial-level attention $\widetilde{m}_c$ is exploited by connection network through the weighted pooling layer, which first multiplies the spatial-level attention with the corresponding output of convolutional layer in the same region, then conducts the pooling operation. By this way, we transfer the spatial-level attention into connection network to emphasize the salient regions with discriminative information.

\subsubsection{Temporal-level attention network}

For a video sequence, different frames contribute to video classification differently. Some frames contain discriminative information, which play the key role for video classification. We propose temporal-level attention network to obtain the discriminative frames for video classification. 
As shown in Fig. \ref{fig:framework}, the temporal-level attention network is composed of an LSTM layer and a softmax layer. This network accepts the output of feature output layer (weighted pooling layer) in connection network as input, and capture temporal contextual information of input video sequence. We deploy two softmax layers followed by two cross-entropy loss layers after the feature output layer and LSTM layer respectively, which drive the joint training of temporal-level network and connection network using the supervision information of video categories. 
By this way, we achieve effective learning of the temporal-level attention network, and the hidden states of LSTM layer can capture temporal evolutions of a particular category in video sequence.

Formally, for an input video sequence ${\bm{x} = [{\bm x}_1, {\bm x}_2, ..., {\bm x}_T]}$, the LSTM maps the input to an output sequence $[{\bm h}_1, {\bm h}_2, ..., {\bm h}_T]$. We stack the hidden states of the LSTM layer as ${\bm H} = [{\bm h}_1, {\bm h}_2, ..., {\bm h}_T] \in \mathbb{R}^{n \times T}$, where every ${\bm h}_i$ encodes the contextual information of the frame in the whole video sequence, as indicated in \cite{liu2017global}. Similar to \cite{lu2016hierarchical}, we calculate the affinity matrix ${\bm C}$ as
\begin{equation}
  {\bm C} = tanh({\bm H}^T {\bm H})
\end{equation}
${\bm C}$ computes the affinity score corresponding to each frame pair in the video sequence, which shows the relevance of each frame pair.
Then we compute the temporal-level attention as column-wise sum of ${\bm C}$:
\begin{equation} 
\label{equ:temporal_attention}
(\gamma_1, \gamma_2, ..., \gamma_T) =  {\bm1^T} {\bm C}
\end{equation}
where ${\bm 1}$ is a vector with all elements to be $1$. $(\gamma_1, \gamma_2, ..., \gamma_T)$ indicates the relevance of every frame respect to the whole video sequence, and the frame with high relevance to the whole sequence is captured as distinguishing frame.

For a video sequence ${\bm{x} = [{\bm x}_1, {\bm x}_2, ..., {\bm x}_T]}$, we denote the output feature of connection network with spatial-level attention as $[\alpha_1, \alpha_2, ..., \alpha_T]$, where $\alpha_i \in \mathbb{R}^{c \times 1}$, $c$ stands for the dimension of feature. Exploiting the temporal-level attention $[\gamma_1, \gamma_2, ..., \gamma_T]$, we can obtain

\begin{equation}
\label{equ:sta_feature}
  \beta_i = \frac {\alpha_i \cdot exp(\gamma_i)} {\sum_{j=1}^T exp(\gamma_j)}, i = {1, 2, ..., T}
\end{equation}

The obtained feature $\beta_i$ contains both discriminative spatial and temporal attention information.
\subsection{Static-motion collaborative model}
This subsection introduces our proposed static-motion collaborative model. We first design a collaborative learning network to mutually boost the representations of static and motion information. Then an adaptively weighted learning approach is proposed to obtain the fusion weights of static and motion streams for each category, which can distinguish different roles of these two streams to improve classification performance.

\subsubsection{Collaborative learning network}
Video naturally contains static and motion information, which are two correlative and complementary aspects to represent the same semantic category. As indicated in \cite{53.feichtenhofer2016convolutional}, separately learning of static and motion information is not enough to fully exploit important clues provided by static and motion information for recognizing video content.
We propose the collaborative learning network, which takes the frame (static) and optical flow (motion) features extracted by spatial-temporal attention network as input, and performs an optimization procedure in an alternate manner to exploit the complementary clues between them. This network is composed of a collaborative learning layer and two softmax layers, where the collaborative learning layer is designed to perform the alternate optimization operations and output optimized features, and the softmax layers are used to generate classification scores of frame and optical flow. The collaborative learning network has a natural symmetry structure between static and motion information, which allows the static features to guide the optimization of motion features, and vice versa. 


Formally, at time $t$, we use the frame features to guide the optimization of optical flow features. We denote the optical flow features as $V^m = [v_1^m, v_2^m, ..., v_N^m]$. By adopting the collaborative learning network, we obtain
\begin{align}
\label{equ:co1}
 & H = tanh(W^m V^m + (W_o^s O^s) {\bm 1}^T) \\
\label{equ:co2}
 & z^m = softmax(W_h^{mT} H) \\
 \label{equ:co3}
 & O^m = \sum z_i^m v_i^m
\end{align}
where ${\bm1}$ stands for the vector with all elements being $1$, $O^s$ stands for the video feature merged from frame features obtained at time $t-1$, $z^m$ stands for the learned optimization coefficients on optical flow features, and $O^m$ stands for the video feature merged from optical flow features, $W^m$, $W_o^s$ and $W_h^m$ are the weight parameters.

At time $t+1$, we use the optical flow features to guide the optimization of frame features. We denote the frame features as $V^s = [v_1^s, v_2^s, ..., v_N^s]$. By adopting the collaborative learning network, we can obtain the optimization coefficients on frame features (denoted as $z^s$), and video feature merged from frame features optimized by $z^s$ (denoted as $O^s$). Above alternate learning procedure achieves the optimization of frame and optical flow features by exploiting the strong complementary clues between static and motion information. The collaborative optimization algorithm is briefly summarized as Algorithm \ref{alg:collaborativelearning}.

\begin{algorithm}
\caption{: Collaborative Learning}
\label{alg:collaborativelearning}
  \begin{algorithmic}[1]
  \REQUIRE
  The features of frame and optical flow extracted from spatial-temporal attention model, $V^s = [v_1^s, v_2^s, ..., v_N^s]$, and $V^m = [v_1^m, v_2^m, ..., v_N^m]$.
  \ENSURE The optimized frame features $V_f^s$ and optical flow features $V_f^m$.
  \STATE Initialize optimization coefficients on frame features as $z^s$ where all of its $N$ elements are set to be 1/$N$.
  \REPEAT
  \STATE Merge the frame features $V^s$ into a single vector as a video feature $O^s = \sum z_i^s v_i^s$.
  \STATE Optimize the optical flow features using $O^s$ by Equation (\ref{equ:co1}) and (\ref{equ:co2}), and obtain the optimization coefficients $z^m$ on the optical flow features.
  \STATE Merge the optical flow features $V^m$ into a single vector as a video feature by Equation (\ref{equ:co3}).
  \STATE Optimize the frame features using $O^m$, and obtain the optimization coefficients $z^s$ on frame features.
  \UNTIL The loss functions converge.
  \RETURN The optimized frame features $V_f^s = {V^s}^T z^s$, and the optimized optical flow features $V_f^m = {V^m}^T z^m$.
  \end{algorithmic}
\end{algorithm}

\subsubsection{Adaptively weighted learning}

As we have obtained the prediction score of each stream (static and motion), we can simply sum up those scores and get the final results. However, static and motion information contribute differently to different video categories. Some categories don't have apparent motion, such as ``Archery'' and ``Smoke'', which should be identified mainly from still frames (static information). While some categories contain obvious motion, and motion clues are important for distinguishing them, such as ``Walk'' and ``Front Crawl''. Therefore, we adaptively learn different fusion weights of static and motion streams for different categories.

Formally, we denote the prediction score of $i$-th training data in $j$-th category as $S_i^j = [{s_{i,1}^j}^T, {s_{i,2}^j}^T]^T \in \mathbb{R}^{2 \times c}$, where $c$ denotes the number of category, $s_{i,m}^j \in \mathbb{R}^{1 \times c}$ stands for the score of $m$-th stream for $i$-th training data in $j$-th category, and we denote the fusion weight for $j$-th category as $W_j = [w_{j,1}, w_{j,2}]$, with the restriction that $\sum_{i=1}^2 w_{j,i} = 1, w_{j,i} > 0$.
The fusion weight for each category is learned separately, and to obtain the weight $W_j$, we define the objective function as:
\begin{equation}
\label{equ:wal_objective}
    arg \max \limits_{W_j} \mathcal{P}_j - \lambda \mathcal{N}_j
\end{equation}
$\mathcal{P}_j$ is defined as:
\begin{equation}
    \mathcal{P}_j = \sum_{i=1}^{n_j} W_j S_i^j J_j
\end{equation}
where $n_j$ stands for the number of training data in $j$-th category. $J_j = [0,...,1,...,0]^T \in \mathbb{R}^{c \times 1}$, with the $j$-th element being $1$, and other elements being $0$. The goal of maximizing $\mathcal{P}_j$ is to maximize the product of $W_j$ and the $j$-th column vector of $S_i^j$. Similarly, we define:
\begin{equation}
    \mathcal{N}_j = \sum_{\{k=1,k \neq j\}}^c \sum_{i=1}^{n_k} W_j S_i^k J_j
\end{equation}
which means minimizing the product of $W_j$ and the $j$-th column vector of $S_i^k$ ($k \neq j$). $\mathcal{P}_j$ and $\mathcal{N}_j$ consider the relationship of positive and negative training data for $W_j$ respectively, and $\lambda$ is the parameter to balance the weight of positive and negative samples. Then the Equation (\ref{equ:wal_objective}) can be transformed to
\begin{eqnarray}
\label{equ:wal_objective_transform}
    \nonumber & arg \max \limits_{W_j} W_j (\sum_{i=1}^{n_j} S_i^j J_j - \lambda \sum_{\{k=1,k \neq j\}}^c \sum_{i=1}^{n_k} S_i^k J_j), \\
    & s.t. \sum_{i=1}^2 w_{j,i} = 1, w_{j,i} > 0
\end{eqnarray}
and the fusion weight can be learned by linear programming easily \cite{116.karmarkar1984new}.

As for the test data, we first calculate and stack the softmax score of each stream, which is denoted as $S_t = [{s_{t,1}}^T, {s_{t,2}}^T]^T \in \mathbb{R}^{2 \times c}$, and the classification result is predicted by
\begin{equation}
    arg \max \limits_{i} W_i S_t J_i
\end{equation}
The final classification results are determined by the highest fusion score.

\section{Experiments}
We conduct experiments on 4 widely-used datasets for video classification, including 3 trimmed video datasets: HMDB51, UCF50, and UCF101, and a large-scale untrimmed video dataset THUMOS14. Our proposed TCLSTA approach is compared with more than 10 state-of-the-art methods to verify its effectiveness.
In the following subsections, we first introduce the 4 datasets briefly, then present the implementation details of our TCLSTA approach, and finally show experimental results and analyses.

\subsection{Datasets}

\begin{itemize}
\renewcommand{\labelitemi}{$\vcenter{\hbox{\scriptsize$\bullet$}}$}
\item {\bf HMDB51} \cite{93.kuehne2011hmdb} dataset provides 3 train-test splits, each of them consists of 6,766 videos, with a fixed frame rate of 30 frames per second (FPS). These clips are labeled with 51 categories of human actions and each video is only labeled with one category. 

\item {\bf UCF50} \cite{94.reddy2013recognizing} dataset consists of 6,618 real-world videos taken from YouTube with a fixed frame rate of 25 FPS, which are labeled with 50 action categories, ranging from general sports to daily life exercises.
These videos are split into 25 groups and videos in the same group may share some common content, such as the same person, similar background or similar viewpoint.

\item {\bf UCF101} \cite{36.soomro2012ucf101} dataset is one of the most popular video classification datasets. It is an extension of UCF50 and consists of 13,320 video clips, which are classified into 101 categories.
These 101 categories can be classified into 5 types (Body motion, Human-human interactions, Human-object interactions, Playing musical instruments and Sports).
The total length of these video clips is over 27 hours. All the videos are collected from YouTube and have a fixed frame rate of 25 FPS with the resolution of 320 $\times$ 240. For the split of training and test sets, we follow the common setting of 3 train-test splits \cite{9.wu2015modeling, 53.feichtenhofer2016convolutional}.

\item {\bf THUMOS14} \cite{THUMOS14} dataset is a large-scale video dataset. It takes UCF101 as its training set and also has background, validation and test sets, with 101 categories. The background set has 2,500 untrimmed long videos. The validation and test set contain 1,010 and 1,584 temporally untrimmed long videos respectively, and they are 178GB in total. Following \cite{wang2017untrimmednets, jain201515}, we use the training set and validation set as training data and evaluate our proposed TCLSTA approach on test set.
\end{itemize}

\begin{figure*}[!t]
\begin{center}
  \includegraphics[width=1.0\textwidth]{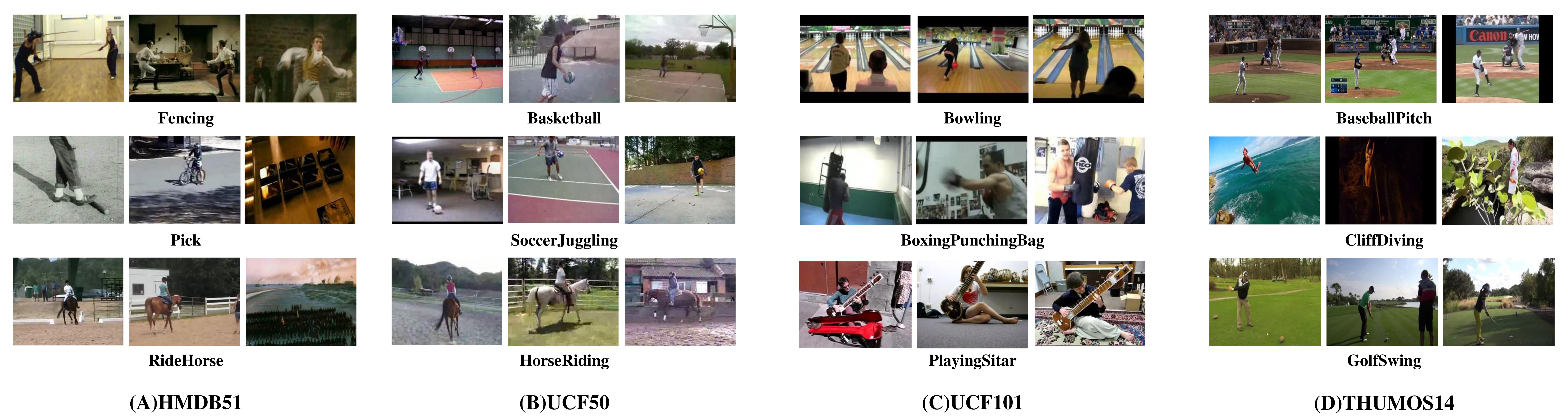}
  \caption{Examples of HMDB51, UCF50, UCF101 and THUMOS14 datasets. These 4 datasets have 51, 50, 101 and 101 categories respectively and we present examples taken from 3 different categories of each dataset.}
  \label{fig:dataset}
\end{center}
\end{figure*}

Examples of these 4 datasets are shown in Fig. \ref{fig:dataset}. Following \cite{2.wang2013action, 53.feichtenhofer2016convolutional}, for UCF50 dataset, we apply the leave-one-group-out cross-validation and report average accuracy over all categories.
In detail, we conduct 25 groups of experiments in UCF50 dataset, where each time we select 24 groups as training set, and the remaining 1 group of videos as test set.
For HMDB51 and UCF101 datasets, we evaluate the results by averaging accuracy over the 3 splits of training and test data.
For THUMOS14 dataset, following \cite{THUMOS14}, mean average precision (MAP) is reported for evaluation.

\subsection{Implementation details}

This subsection introduces the implementation details including network structures and training details in the experiments.

\subsubsection{Spatial-temporal attention model} We adopt the same network architectures on both two streams of frame and optical flow. For the optical flow stream, we pre-compute the optical flow using TVL1 method \cite{zach2007duality} and store the flow fields as JPEG images. Following \cite{15.simonyan2014two}, we stack flow fields of every $L=10$ consecutive frames into a $2L$-channel image, which takes the horizontal and vertical components of flow fields as channels.  
The implementation details of three component networks of spatial-temporal attention network are presented as follows:

{\bf Connection network:}
The connection network is constructed based on ResNet-50\cite{95.he2016deep}, and Table \ref{tab:connection network} presents its detailed architecture.
Each column shows the building block(s) of layer(s) with the same output size. The convolutional filters are shown as $(W\times H, C)$, denoting {\bf W}idth, {\bf H}eight and number of feature {\bf C}hannels, and the pooling sizes are shown as $(W\times H)$. For the columns that present multiple blocks, the numbers of blocks stacked are presented to the right of brackets. The ``Input size'' and ``Output size'' are shown as $(W\times H \times C)$. The input size of \emph{conv1} is $224\times 224\times 3$ for frame and $224\times 224\times 20$ for optical flow. 
The weighted pooling layer is also a \textbf{\emph{feature output layer}} as its output can be treated as a feature of size $1\times 1\times 2048$. Both frame feature and optical flow feature are extracted from the feature output layer, and the dimensions of them are both 2048. Following the weighted pooling layer, a softmax layer is deployed for classification as shown in Fig. \ref{fig:framework}.
In the training phase, we apply the ResNet-50 model pre-trained on ImageNet dataset and set the number of hidden units in softmax (classification) layer as the category number of the corresponding dataset.
The ResNet-50 used in our approach can be replaced by other CNN structures, such as AlexNet \cite{3.krizhevsky2012imagenet}, VGGNet \cite{simonyan15verydeep}, and GoogLeNet \cite{szegedy2015going}, etc.
In the experiments, we choose ResNet-50 for implementation due to its wide applications and performance advantage in various tasks of computer vision as reported in previous works as \cite{95.he2016deep}.

{\bf Spatial-level attention network:}
As shown in Table \ref{tab:spatial network}, the spatial-level attention network has the same architecture as connection network from \emph{conv1} layer to \emph{conv5\_x} layers, and shares the weights on these layers. Following \cite{52.zhou2016learning}, a convolutional layer (\emph{CAM\_conv} layer) with $1024$ filters and a GAP layer are sequentially deployed after the shared convolutional layers. Then we deploy a softmax layer following the GAP layer with $N$ hidden units, where $N$ is set to be the category number of the corresponding dataset. 

{\bf Temporal-level attention network:}
The temporal-level attention network is composed of a LSTM layer and a softmax layer, as shown in Table \ref{tab:temporal network}. The LSTM layer has $512$ LSTM units and the softmax layer has $N$ hidden units, where $N$ is set to be the category number of the corresponding dataset.

For the training of spatial-temporal attention model, we crop the input frames and flow images to the size of $224 \times 224$. Then we use mini-batch stochastic gradient descent (SGD) to optimize the neural networks with each mini-batch containing 64 frames or $2L$-channel flow images. We set the learning rate to be 0.001 initially, and reduce it by a factor of 10 after the validation accuracy saturates. The weight decay is set to be 0.0001 and the momentum to be 0.9. For HMDB51, UCF50, and UCF101 datasets, we train the networks for 30K iterations on frame stream, and for 60K iterations on optical flow stream, because the networks on optical flow stream have lower convergence rates. For THUMOS14 dataset, we train the networks for 60K iterations on frame stream, and for 120K iterations on optical flow stream, since THUMOS14 dataset has much more frames and flow images for training.
We apply three cross-entropy loss functions (three loss layers) after the softmax layers in connection network, spatial-level attention network and temporal-level attention network, which drive the joint learning of the whole spatial-temporal attention model with the same supervised information.

\subsubsection{Static-motion collaborative model}
The static-motion collaborative model is composed of a collaborative learning network and an adaptively weighted learning model. {\bf Collaborative learning network} consists of a collaborative learning layer and two softmax layers, as shown in Fig. \ref{fig:framework}. In detail, collaborative learning layer accepts static and motion features as inputs, and optimizes them by alternate optimization procedure as illustrated in Algorithm \ref{alg:collaborativelearning}. Similar to spatial-level attention model, the two softmax layers are designed with $N$ hidden units, where $N$ denotes the number of classes in corresponding dataset. Finally, cross-entropy loss is adopted to optimize the collaborative learning network. For {\bf adaptively weighted learning model}, we set the parameter $\lambda$ in Equation (\ref{equ:wal_objective_transform}) to be $5 \times 10^{-3}$, which is selected by the cross-validation model.
\begin{table*}[!t]
\newcommand{\tabincell}[2]{\begin{tabular}{@{}#1@{}}#2\end{tabular}}
\centering
\caption{Architecture of connection network. Building blocks are shown in brackets, with the numbers of blocks stacked. Down-sampling is performed by conv3\_1, conv4\_1, and conv5\_1 with a stride of 2.}
\label{tab:connection network}
\scalebox{0.89}{
\begin{tabular}{|c|c|c|c|c|c|c|c|}
\hline
Layers  & conv1 & pool1 & conv2\_x & conv3\_x & conv4\_x & conv5\_x & \tabincell{c}{Weighted pooling\\(Feature output)}\\
\hline
\tabincell{c}{Input size} & \tabincell{c}{$224\times224\times3$\\($224\times224\times20$)} & $112\times112\times64$ & $56\times56\times64$ & $56\times56\times256$ & $28\times28\times512$ & $14\times14\times1024$ & $7\times7\times2048$ \\
\hline
\multirow{4}{*}{Blocks} & 
\multirow{4}{*}{$\begin{matrix}7\times7,64\\\text{stride 2}\end{matrix}$} & 
\multirow{4}{*}{$\begin{matrix}3\times3\\\text{max pool}\\\text{stride 2}\end{matrix}$} &
\multirow{4}{*}{$\left[\begin{matrix}1\times1,64\\3\times3,64\\1\times1,256\end{matrix}\right]\times3$} &
\multirow{4}{*}{$\left[\begin{matrix}1\times1,128\\3\times3,128\\1\times1,512\end{matrix}\right]\times4$} &
\multirow{4}{*}{$\left[\begin{matrix}1\times1,256\\3\times3,256\\1\times1,1024\end{matrix}\right]\times6$} &
\multirow{4}{*}{$\left[\begin{matrix}1\times1,512\\3\times3,512\\1\times1,2048\end{matrix}\right]\times3$} &
\multirow{4}{*}{$\begin{matrix}7\times7\\\text{weighted pool}\end{matrix}$} \\
&&&&&&&\\
&&&&&&&\\
&&&&&&&\\
\hline
\tabincell{c}{Output size} & $112\times112\times64$ & $56\times56\times64$ & $56\times56\times256$ & $28\times28\times512$ & $14\times14\times1024$ & $7\times7\times2048$ & $1\times1\times2048$ \\
\hline
\end{tabular}
}
\end{table*}

\subsection{Comparisons with state-of-the-art methods}

This subsection presents the experimental results and analyses of our TCLSTA approach on 3 trimmed video datasets and 1 untrimmed video dataset compared with state-of-the-art methods. All the compared results are presented in Table \ref{tab:trimmed}.
From Table \ref{tab:trimmed}, for HMDB51 dataset, we can see early works \cite{112.cai2014multi, 114.wang2013dense} choose hand-crafted features for video representation, and the performances are limited and much lower than our TCLSTA approach.
Some works such as \cite{15.simonyan2014two} employ two types of CNNs to model static and motion information, and achieve better performances than the traditional video classification methods \cite{112.cai2014multi, 114.wang2013dense}. However, the improvement is limited due to the simple fusion strategy.
Besides, some methods \cite{53.feichtenhofer2016convolutional, 103.feichtenhofer2016spatiotemporal} apply more complex fusion methods for combining static and motion information, and achieve better results than \cite{15.simonyan2014two}. But on the one hand, they ignore the strong complementarity between static and motion information coexisting in video, and the features of static and motion information are learned separately. On the other hand, they ignore the spatial-temporal attention, thus the features learned by these methods are not discriminative enough.
Our TCLSTA achieves the best result among these state-of-the-art methods, bringing a promotion of 2.3\% than the highest result of compared methods.
It is for the reason that our TCLSTA not only exploits the strong complementarity between static and motion information to guide and mutually boost the learning of static and motion streams, but also jointly learns the discriminative features of video by our proposed spatial-temporal attention model.

\begin{table}[!t]
\newcommand{\tabincell}[2]{\begin{tabular}{@{}#1@{}}#2\end{tabular}}
\centering
\caption{Architecture of spatial-level attention network}
\label{tab:spatial network}
\scalebox{1.0}{
\begin{tabular}{|c|c|c|c|}
\hline
Layers  & conv1 - conv5\_x & CAM\_conv & GAP \\
\hline
\tabincell{c}{Input size} & \tabincell{c}{$224\times224\times3$\\($224\times224\times20$)} & $7\times7\times2048$ & $7\times7\times1024$ \\
\hline
Blocks & 
\tabincell{c}{the same as \\connection network} & 
$3\times3,1024$ &
\tabincell{c}{$7\times7$\\average pool} \\
\hline
\tabincell{c}{Output size} & $7\times7\times2048$ & $7\times7\times1024$ & $1\times1\times1024$ \\
\hline
\end{tabular}
}
\end{table}

\begin{table}[!t]
\newcommand{\tabincell}[2]{\begin{tabular}{@{}#1@{}}#2\end{tabular}}
\centering
\caption{Architecture of temporal-level attention network}
\label{tab:temporal network}
\scalebox{1.0}{
\begin{tabular}{|p{2.5cm}<{\centering}|p{2cm}<{\centering}|}
\hline
Layers  & LSTM \\
\hline
Input size & $2048$  \\
\hline
Hidden units & $512$ \\
\hline
Output size & $512$ \\
\hline
\end{tabular}
}
\end{table}

\begin{figure}[!t]
  \centering
  \includegraphics[width=0.5\textwidth]{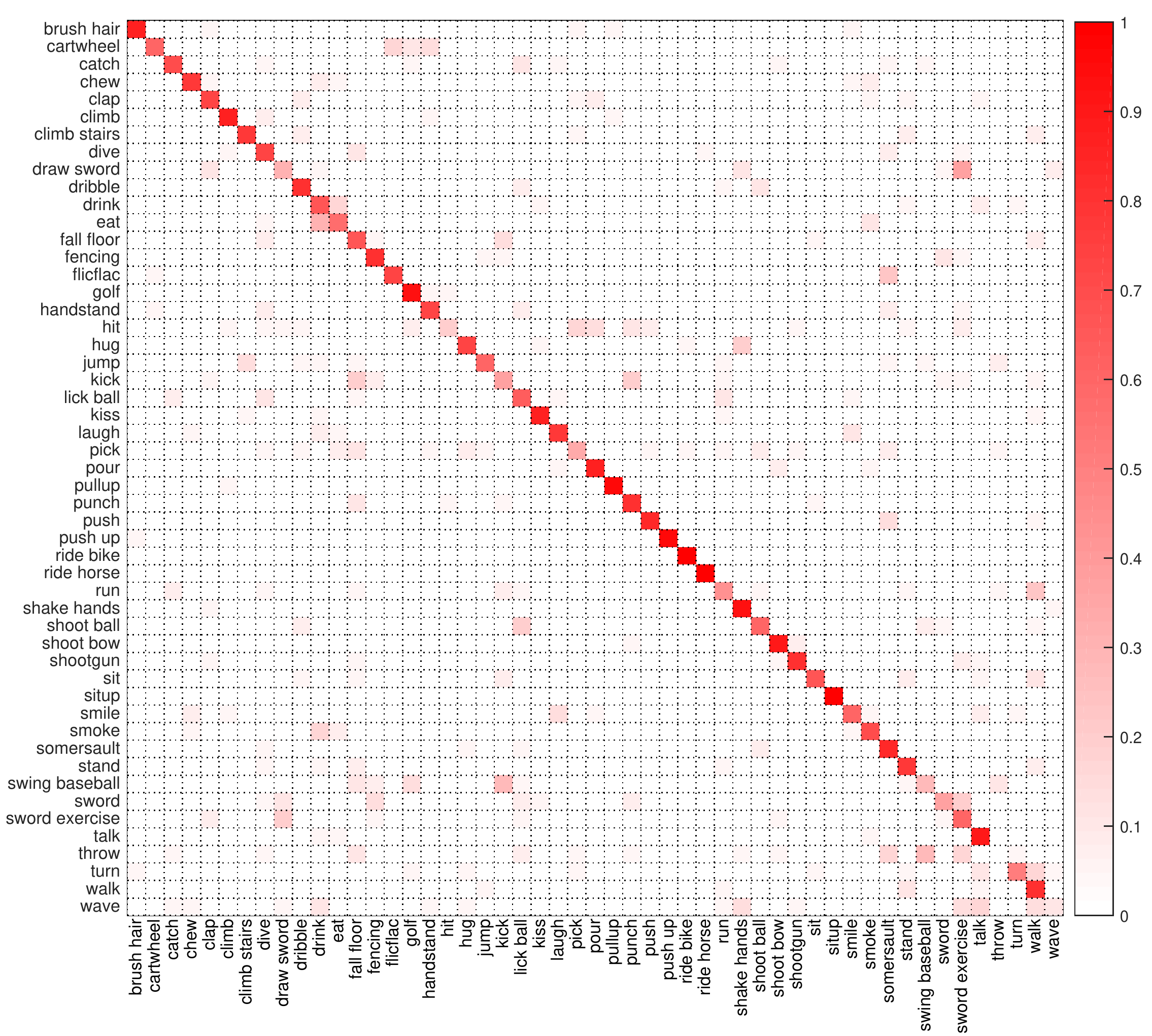}
  \caption{The confusion matrix on HMDB51 dataset. The column and row of the matrix represent the categories in HMDB51 dataset.}
  \label{fig:cmhmdb51}
\end{figure}

\begin{table*}[!t]
\setlength{\abovecaptionskip}{-5pt}
\caption{Experimental results compared with state-of-the-arts on HMDB51, UCF50, UCF101 and THUMOS14 datasets.  
The evaluation metric for HMDB51, UCF50 and UCF101 datasets is accuracy, and the evaluation metric for THUMOS14 dataset is MAP.}
\label{tab:trimmed}
\begin{center}
\scalebox{1.0}{
	\begin{tabular}{|p{3cm}|c|p{3cm}|c|p{3cm}|c|p{3cm}|c|}
	\hline
	\multicolumn{2}{|c|}{HMDB51}&\multicolumn{2}{c|}{UCF50}&\multicolumn{2}{c|}{UCF101}&\multicolumn{2}{c|}{THUMOS14}\\
	\hline
	\hline
	\textbf{Ours} & \textbf{0.687} & \textbf{Ours} &  \textbf{0.957} & \textbf{Ours} & \textbf{0.940} & \textbf{Ours} & \textbf{0.847}\\
	\hline
	\hline
	Feichtenhofer et al. \cite{103.feichtenhofer2016spatiotemporal} & 0.664 & Lan et al. \cite{98.lan2015beyond} & 0.944            & Feichtenhofer et al. \cite{103.feichtenhofer2016spatiotemporal} & 0.934  & Wang et al. \cite{wang2017untrimmednets} & 0.822 \\

	Wang et al. \cite{108.wang2016deep} & 0.659                             & Chen et al. \cite{101.chen2010long} & 0.930           & Zhu et al. \cite{105.zhu2016key}                                         & 0.931 & Wang et al. \cite{wang2016temporal} & 0.785 \\

	Feichtenhofer et al. \cite{53.feichtenhofer2016convolutional} & 0.654   & Yang et al. \cite{yang2017deep} & 0.924               & Feichtenhofer et al. \cite{53.feichtenhofer2016convolutional} & 0.925    &Jain et al. \cite{jain201515} & 0.716 \\

	Bilen et al. \cite{109.bilen2016dynamic} & 0.652                        & Peng et al. \cite{100.peng2016bag} & 0.923            & Li et al. \cite{104.li2016vlad3} & 0.922                                 &Simonyan et al. \cite{15.simonyan2014two} & 0.661 \\

	Fernando et al. \cite{110.fernando2015modeling} & 0.637                 & Wang et al. \cite{99.wang2016robust} & 0.917          & Wu et al. \cite{9.wu2015modeling} & 0.913                                & Varol et al. \cite{varol2015efficient} & 0.634 \\

	Zhu et al. \cite{105.zhu2016key} & 0.633                                & Wang et al. \cite{2.wang2013action} & 0.912           & Lan et al. \cite{98.lan2015beyond} & 0.891                               & Wang et al. \cite{2.wang2013action} & 0.631 \\

	Simonyan et al. \cite{15.simonyan2014two} & 0.594                       & Oneata et al. \cite{96.oneata2013action} & 0.900      & Yue et al. \cite{23.yue2015beyond} & 0.882                               & Zhang et al. \cite{zhang2016real} & 0.615 \\

	Wang et al. \cite{2.wang2013action} & 0.572                             & Ciptadi et al. \cite{113.ciptadi2014movement} & 0.900 & Sun et al. \cite{56.sun2015human} & 0.881                                & &\\
	Wu et al. \cite{111.wu2014towards} & 0.564                              & Narayan et al. \cite{97.narayan2014cause} & 0.894     & Simonyan et al. \cite{15.simonyan2014two} & 0.880                        & &\\
	Cai et al. \cite{112.cai2014multi} & 0.559                              & &                                                     & Wang et al. \cite{2.wang2013action} & 0.859                              & &\\
	Wang et al. \cite{114.wang2013dense} & 0.466                            & &                                                     & Karpathy et al. \cite{7.karpathy2014large} & 0.660                       & &\\
	\hline
	\end{tabular}
}
\end{center}
\end{table*}

The comparison results on UCF50 and UCF101 datasets are also shown in Table \ref{tab:trimmed}. The trends of results on these two datasets are similar with HMDB51, and our TCLSTA approach achieves the best results (0.957 and 0.940 respectively) among state-of-the-art methods, and brings 1.3\% and 0.6\% improvements respectively.

We also show the confusion matrix on HMDB51 dataset as Fig. \ref{fig:cmhmdb51}, in which the columns and rows represent the categories in HMDB51 dataset.
The confusion matrix is a probability matrix, where the element at $i$-th row and $j$-th column represents the probability of predicting $i$-th category to be $j$-th category. And from Fig. \ref{fig:cmhmdb51} we can see that our TCLSTA approach is of good performance for most categories.

\begin{figure}[h!]
  \begin{center}
  \includegraphics[width=0.5\textwidth]{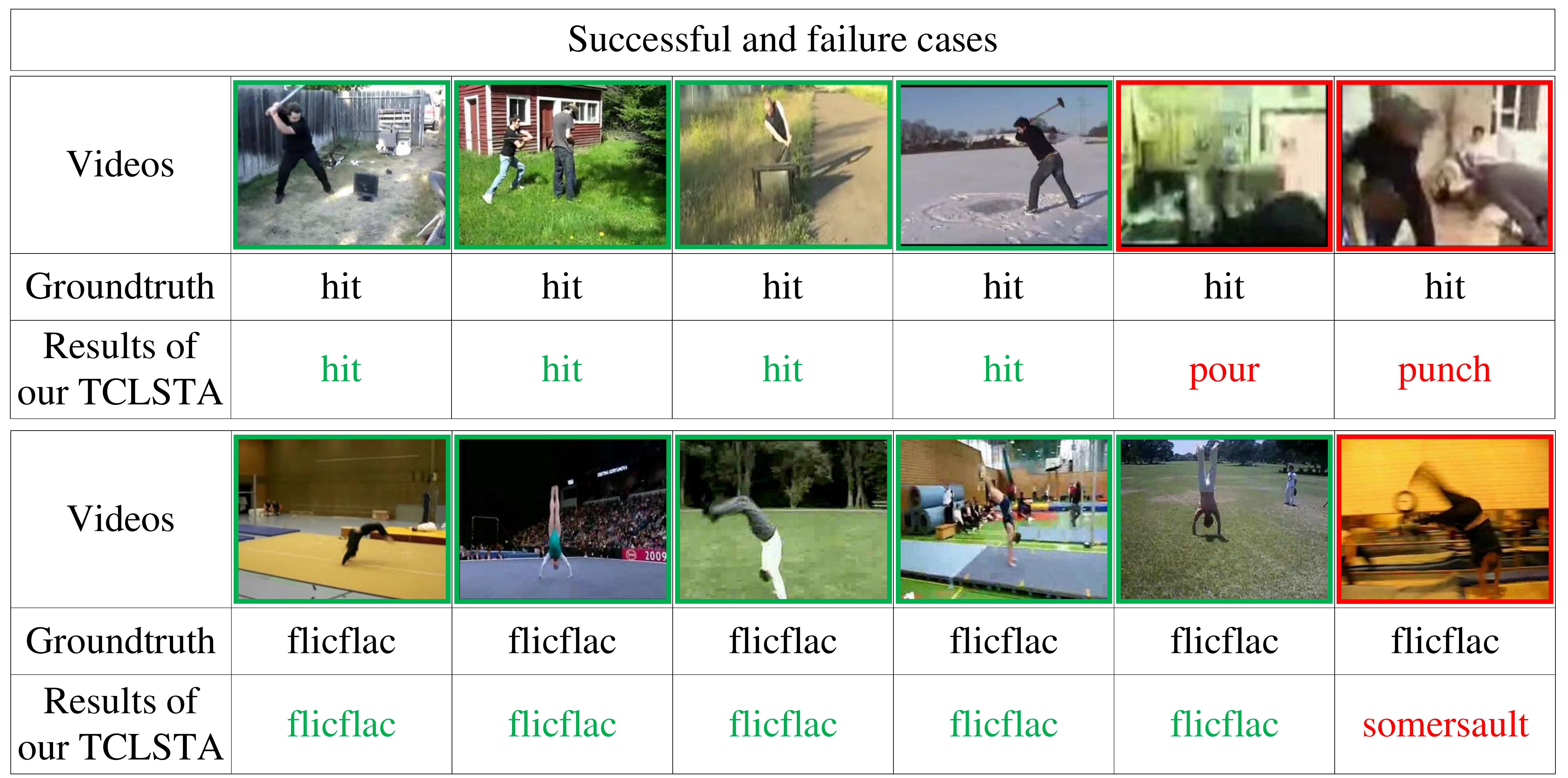}
  \caption{Some classification results of successful (denoted as green boxes) and failure (denoted as red boxes) cases on HMDB51 dataset. Our proposed TCLSTA approach fails on some examples due to low resolution and large camera motion, which makes these examples very difficult to recognize.}
  \label{fig:hmdb51_error}
  \end{center}
\end{figure}

Compared with results on UCF50 and UCF101 datasets, our TCLSTA approach achieves relatively low accuracy on HMDB51 dataset. This is because HMDB51 dataset is very challenging due to the large camera motion, various viewpoints, and low clip quality, which result in high intra-class variations, so the performances of all compared methods and our TCLSTA approach decrease on this dataset. Fig. \ref{fig:hmdb51_error} presents some examples of successful and failure cases of TCLSTA approach on HMDB51 dataset. From Fig. \ref{fig:hmdb51_error} we can see that the videos have low resolution or intensive camera motion, which are very challenging to be classified accurately. However, our proposed TCLSTA approach keeps the best accuracy even on this challenging dataset compared with 11 state-of-the-art methods, which shows its effectiveness and generality.

The last column of Table \ref{tab:trimmed} shows the comparison results on the untrimmed video dataset THUMOS14,  between our proposed TCLSTA approach and the state-of-the-art methods. It is noted that results of \cite{15.simonyan2014two, 2.wang2013action, zhang2016real,wang2016temporal} are cited from \cite{wang2017untrimmednets}. THUMOS14 is a very challenging dataset, because it has large-scale temporally untrimmed video data that contains long-period content without any instance of pre-defined categories. From Table \ref{tab:trimmed}, we can see that despite the great challenges of this dataset, our proposed TCLSTA still improves the classification performance, which fully demonstrates that our proposed TCLSTA approach can also effectively handle the untrimmed videos.

\subsection{Performances of components in our TCLSTA approach}

To further evaluate each component of proposed TCLSTA approach, we conduct detailed experiments from the following two aspects.

\subsubsection{Effectiveness of spatial-temporal attention model}

\begin{table}[t]
\setlength{\abovecaptionskip}{5pt}%
\setlength{\belowcaptionskip}{10pt}%
\caption{Experimental results compared with different levels of attention and their combinations. ``TH14'' denotes THUMOS14 dataset. Results of the four experiments in ``Two-stream" setting is obtained by averaging the softmax score of both streams.}
\label{tab:attention}
\begin{center}
\scalebox{1.0}{
\begin{tabular}{|p{2.5cm}|c|c|c|c|}
\hline
      Methods    &     HMDB51  &      UCF50 &     UCF101      &TH14\\
\hline \hline
Frame         &     0.514      &     0.904  &     0.839       &0.721 \\
Frame + SA    &     0.527      &     0.912  &     0.847       &0.744 \\
Frame + TA    &     0.535      &     0.920  &     0.850       &0.729 \\
Frame + STA   &     0.548      &     0.926  &     0.859       &0.763 \\
\hline \hline
Optical flow &     0.529&     0.900          &     0.810       &0.665  \\
Optical flow + SA &     0.547&   0.910       &     0.827       &0.688 \\
Optical flow + TA &     0.565&        0.913  &     0.840       &0.671 \\
Optical flow + STA &     0.592&     0.919    &     0.859       &0.702 \\
\hline \hline
Two-stream &     0.626 &     0.934          &     0.917       &0.776 \\
Two-stream + SA  &     0.641&     0.939     &     0.923       &0.800 \\
Two-stream + TA &     0.652 &     0.941     &     0.926       &0.796 \\
Two-stream + STA &     0.676 &    0.948     &     0.928       &0.831 \\
\hline
\end{tabular}  }
\end{center}
\end{table}

Our TCLSTA approach involves two streams and two kinds of attention. We denote the two streams and the combination of them as ``Frame'', ``Optical flow'' and ``Two-stream'' respectively. Similarly, we denote spatial-level attention, temporal-level attention and their combination, spatial-temporal attention, as ``SA'', ``TA'' and ``STA'' respectively. The experimental results with the above components are shown in Table \ref{tab:attention}, from which we observe that:

\begin{itemize}
\renewcommand{\labelitemi}{$\vcenter{\hbox{\scriptsize$\bullet$}}$}
  \item Both spatial and temporal attention are helpful for improving the classification accuracy by highlighting the discriminative ``parts'' of frame and optical flow, in which spatial-level attention helps to highlight the salient regions, while the temporal-level attention helps to highlight the discriminative frames in a video sequence. Taking HMDB51 and UCF50 datasets for examples, on frame stream, comparing with the results of ``Frame'', spatial-level attention and temporal-level attention improve the classification accuracy by promotions of 1.3\% and 2.1\% respectively on HMDB51 dataset as well as promotions of 0.8\% and 1.6\% respectively on UCF50 dataset. On optical flow stream, comparing with the results of ``Optical flow'', we can see spatial-level attention and temporal-level attention achieve the promotions of 1.8\%, 3.6\% on HMDB51 dataset as well as promotions of 1.0\%, 1.3\% on UCF50 dataset. And on the setting of ``Two-stream'', we can also see the accuracy improvement achieved by spatial-level attention and temporal-level attention, e.g. 65.2\% and 64.1\% vs. 62.6\% on HMDB51 dataset as well as 93.9\% and 94.1\% vs. 93.4\% on UCF50 dataset. Similar improvements can also be observed on UCF101 and THUMOS14 datasets, which validate the effectiveness of spatial-level and temporal-level attention for promoting the performance of video classification.
  \item Combining the spatial-level attention and temporal-level attention further improves the accuracies than exploiting only one type of attention. Taking the ``Two-stream'' setting for illustration, compared to the results of only with spatial-level attention, spatial-temporal attention achieves promotions of 3.5\%, 0.9\%, 0.5\% and 3.1\% on 4 datasets respectively. Compared to the results of only with temporal-level attention, it achieves promotions of 2.4\%, 0.7\%, 0.2\% and 3.5\% on 4 datasets respectively. And comparing with the results without any attention on 3 settings of ``Frame'', ``Optical flow'' and ``Two-stream'', spatial-temporal attention achieves promotions of 3.4\%, 6.3\%, 5.0\% on HMDB51 dataset, 2.2\%, 1.9\%, 1.4\% on UCF50 dataset, and similar promotions can be seen on UCF101 and THUMOS14 datasets. It shows the effectiveness of our spatial-temporal attention model on jointly learning the spatial-level and temporal-level attention, because the spatial-temporal attention is coexistent in frame and optical flow, thus should be jointly modelled as spatial and temporal evolutions of videos. 
\end{itemize}

\begin{figure}[!t]
  \centering
  \includegraphics[width=0.49\textwidth]{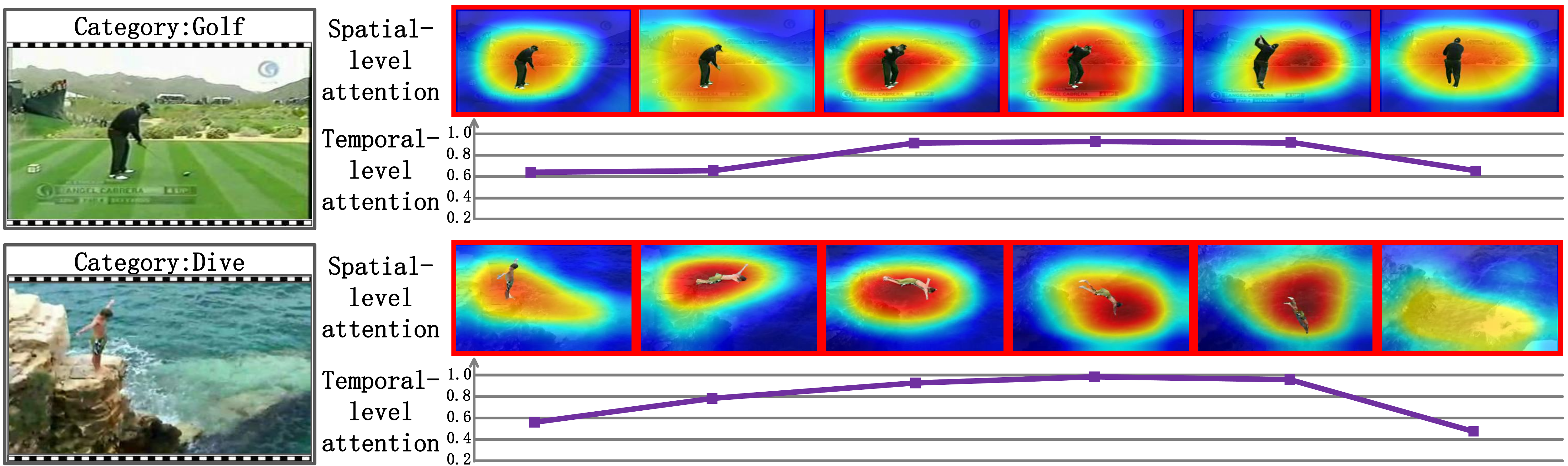}
  \caption{Visualization of spatial-temporal attention on examples of HMDB51 dataset. The heatmaps indicate the importance of different regions in frame, and the line charts indicate the importance of different frames.}
  \label{fig:heatmap_hmdb}
\end{figure}

Fig. \ref{fig:heatmap_hmdb} shows the visualization of spatial-temporal attention detected by our proposed approach on some examples in HMDB51 dataset, using heatmaps and line charts. The heatmaps indicate the importance of different regions in frame, and the line charts indicate the importance of different frames. From Fig. \ref{fig:heatmap_hmdb}, we can see that spatial-temporal attention model in our proposed TCLSTA approach captures the discriminative frames and its salient regions accurately.

\subsubsection{Effectiveness of collaborative learning model}

\begin{table}[t]
\setlength{\abovecaptionskip}{-5pt}
\caption{Experimental results of static-motion collaborative model. ``TH14'' denotes THUMOS14 dataset. Results of ``Two-stream + STA" and ``Two-stream + STA + CLN" are obtained by late fusion.}
\label{tab:collaborative}
\begin{center}
  \scalebox{0.94}{
\begin{tabular}{|p{3.6cm}|c|c|c|c|}
\hline
    Methods   &     HMDB51    &      UCF50        &     UCF101  & TH14\\
\hline
\hline
Two-stream+STA &      0.676&     0.948          &      0.928  & 0.831 \\
Two-stream+STA+CLN &      0.679&     0.951    &      0.932  & 0.836 \\
Two-stream+STA+AWL &      0.682  &     0.953  &      0.936  & 0.841 \\
Two-stream+STA+CLN+AWL  &  0.687&   0.957   &      0.940  & 0.847 \\
\hline
\end{tabular}
}
\end{center}
\end{table}

\begin{table}[t]
\setlength{\abovecaptionskip}{-5pt}%
\setlength{\belowcaptionskip}{10pt}%
\caption{Experimental results compared with different fusion methods. ``TH14'' denotes THUMOS14 dataset.}
\label{tab:fusion}
\begin{center}
  \scalebox{0.94}{
\begin{tabular}{|p{3.6cm}|c|c|c|c|}
\hline
     Methods     &     HMDB51  &      UCF50 &     UCF101                          &TH14\\
\hline
\hline
Two-stream + STA + Late &      0.676      &     0.948       &      0.928   & 0.831 \\
Two-stream + STA + Early &      0.679      &     0.950       &      0.931  & 0.834 \\
Two-stream + STA + MKL  &    0.680        &      0.951      &    0.934      & 0.836 \\
\hline
\hline
Two-stream + STA + AWL  &     0.682   &     0.953      &     0.936                & 0.841 \\
\hline
\end{tabular}
}
\end{center}
\end{table}

\begin{table}[t]
\setlength{\abovecaptionskip}{5pt}%
\setlength{\belowcaptionskip}{10pt}%
\caption{Experimental results about efficiency evaluation on HMDB51 dataset. }
\label{tab:efficiency}
\begin{center}
  \scalebox{1.1}{
  \begin{tabular}{|p{3.5cm}|c|}

  \hline
  Methods & Frames per second \\
  \hline
  \hline
  \textbf{Ours} & \textbf{89.6} \\
  \hline
  \hline
  Feichtenhofer et al. \cite{103.feichtenhofer2016spatiotemporal} & 49.5 \\
  Wang et al. \cite{108.wang2016deep}                             & 26.3 \\
  Feichtenhofer et al. \cite{53.feichtenhofer2016convolutional}   & 33.2 \\
  Bilen et al. \cite{109.bilen2016dynamic}                        & 37.0 \\
  Zhu et al. \cite{105.zhu2016key}                                & 96.7 \\
  Simonyan et al. \cite{15.simonyan2014two}                       & 99.7 \\
  Wang et al. \cite{2.wang2013action}                             & 37.9 \\
  Wang et al. \cite{114.wang2013dense}                            & 48.5 \\
  \hline
  \end{tabular}
}
\end{center}
\end{table}

We further conduct experiments to demonstrate the effectiveness of proposed corroborative learning network and adaptively weighted learning. In Table \ref{tab:collaborative}, ``Two-stream + STA" refers to applying spatial-temporal attention on both streams, then fusing by late fusion. ``CLN" refers to the collaborative learning network, and ``AWL" refers to adaptively weighted learning.

From Table \ref{tab:attention}, comparing the results of 3 settings: ``Frame", ``Optical flow" and ``Two-stream", it is easy to observe that frame and optical flow are complementary in video classification, for the reason that these two streams express the static and motion information respectively, which are two important aspects for representing video information.
Compared with the results of ``Two-stream + STA" in Table \ref{tab:collaborative}, which is without collaborative learning model, ``Two-stream + STA + CLN" achieves better classification accuracy and MAP score. It shows that our proposed collaborative learning network can boost the learning of frame and optical streams mutually, and exploit the correlation between them, thus further improves video classification accuracy.

Compared with the results of ``Two-stream + STA" in Table \ref{tab:collaborative}, ``Two-stream + STA + AWL" also achieves higher classification accuracy, which shows that it is helpful to adaptively learn the fusion weights of two streams. ``Two-stream + STA + CLN + AWL" achieves the best accuracy among all the 4 settings, which shows the collaborative learning network and adaptively weighted learning can reinforce each other.

Furthermore, for validating the effectiveness of our adaptively weighted learning method, we conduct more comparison experiments with different fusion strategies, namely early fusion, late fusion and MKL fusion \cite{117.kloft2011lp}. The brief introductions of these 4 fusion methods are as follows, and the comparison results are shown in Table \ref{tab:fusion}.

\begin{itemize}
\renewcommand{\labelitemi}{$\vcenter{\hbox{\scriptsize$\bullet$}}$}
 \item {\bf Late fusion}: Averaging the prediction scores of static and motion streams (denoted as ``Late'' in Table \ref{tab:fusion}).
 \item {\bf Early fusion}: Concatenating static and motion features and training a SVM for final video classification (denoted as ``Early'' in Table \ref{tab:fusion}).
 \item {\bf MKL fusion}: Using scores as features and combining different kernels by the LP-norm MKL algorithm \cite{117.kloft2011lp} (denoted as ``MKL'' in Table \ref{tab:fusion}).
 \item {\bf Our adaptively weighted learning}: Fusing the scores by the adaptively weighted learning method (denoted as ``AWL'' in Table \ref{tab:fusion}).
\end{itemize}

Table \ref{tab:fusion} shows that the trends of results for these fusion methods are similar among all the 4 datasets. In detail, late fusion and early fusion achieve relatively lower accuracies, as they cannot distinguish different roles of static and motion information for different categories. MKL fusion gets slightly better results than early fusion and late fusion, while its performance gain is not often observed to be significant, as indicated by \cite{Vedaldi2009multiple}. Our adaptively weighted learning method learns specific fusion weight for each category in an adaptive fashion and achieves the best accuracies on all 4 datasets, because it distinguishes different contributions of static and motion information to different semantic categories.

\subsection{Evaluation on efficiency}
For evaluating the efficiency of our proposed TCLSTA approach, we test the running speed in the test procedure on HMDB51 dataset. Table \ref{tab:efficiency} shows the comparison results with state-of-the-art methods. All the results are obtained on a PC that has an Intel i7-5930K CPU, 64GB RAM and a TITAN X GPU with 12GB memory. 
The compared methods are cited from Table \ref{tab:trimmed}, which include methods based on deep networks \cite{103.feichtenhofer2016spatiotemporal,108.wang2016deep,53.feichtenhofer2016convolutional,109.bilen2016dynamic,105.zhu2016key,15.simonyan2014two} and methods based on hand-crafted features \cite{2.wang2013action,114.wang2013dense,110.fernando2015modeling,111.wu2014towards,112.cai2014multi}. It is noted that, the test procedure of methods based on hand-crafted features includes extracting local features, feature encoding and classifying, where the process of extracting local features takes most of the computation cost. So we report the running speed of extracting local features for simplicity on methods based on hand-crafted features. We take \cite{2.wang2013action} and \cite{114.wang2013dense} for representative of methods based on hand-crafted features, since all the other methods \cite{110.fernando2015modeling,111.wu2014towards,112.cai2014multi} utilize dense trajectory features as \cite{114.wang2013dense} or improved dense trajectory features as \cite{2.wang2013action}, so that the efficiency results of \cite{2.wang2013action} and \cite{114.wang2013dense} can relatively accurately show the efficiency of the methods based on hand-crafted features.
From Table \ref{tab:efficiency}, we can see that our proposed TCLSTA approach outperforms all the compared methods except \cite{15.simonyan2014two} and \cite{105.zhu2016key}, both of which utilize relatively shallow networks.
With the cost of slightly lower efficiency than \cite{15.simonyan2014two} and \cite{105.zhu2016key}, our proposed TCLSTA approach clearly outperforms them on accuracy, as shown in Table \ref{tab:trimmed}. 

\section{Conclusions}

This paper has proposed the two-stream collaborative learning with spatial-temporal attention approach (TCLSTA) for video classification, which consists of spatial-temporal attention model and static-motion collaborative model.
Spatial-temporal attention model adopts a spatial-level attention network to emphasize the salient regions of frame, and uses a temporal-level attention network to exploit the discriminative frames in video. Both of them are jointly optimized and mutually boosted to exploit discriminative features.
Static-motion collaborative model employs the discriminative static and motion features extracted by spatial-temporal attention model, to mutually boost the representation and optimize the combining weight of frame and optical flow for video classification.
Experiments on 4 widely-used video classification datasets show that our TCLSTA approach achieves the best performance compared with more than 10 state-of-the-art methods.

The future works lie in two aspects:
First, we will focus on exploiting better spatial-temporal attention, and learning more discriminative static-motion representation.
Second, we will also attempt to apply unsupervised learning into our work, which can make full use of large amount of unlabeled videos on the Internet. Both of them will be jointly employed to further improve the performance of video classification.

\ifCLASSOPTIONcaptionsoff
  \newpage
\fi

\bibliographystyle{IEEEtran}
\bibliography{cite}

\begin{thebibliography}{10}
\providecommand{\url}[1]{#1}
\csname url@samestyle\endcsname
\providecommand{\newblock}{\relax}
\providecommand{\bibinfo}[2]{#2}
\providecommand{\BIBentrySTDinterwordspacing}{\spaceskip=0pt\relax}
\providecommand{\BIBentryALTinterwordstretchfactor}{4}
\providecommand{\BIBentryALTinterwordspacing}{\spaceskip=\fontdimen2\font plus
\BIBentryALTinterwordstretchfactor\fontdimen3\font minus
  \fontdimen4\font\relax}
\providecommand{\BIBforeignlanguage}[2]{{%
\expandafter\ifx\csname l@#1\endcsname\relax
\typeout{** WARNING: IEEEtran.bst: No hyphenation pattern has been}%
\typeout{** loaded for the language `#1'. Using the pattern for}%
\typeout{** the default language instead.}%
\else
\language=\csname l@#1\endcsname
\fi
#2}}
\providecommand{\BIBdecl}{\relax}
\BIBdecl

\bibitem{15.simonyan2014two}
K.~Simonyan and A.~Zisserman, ``Two-stream convolutional networks for action
  recognition in videos,'' in \emph{Advances in Neural Information Processing
  Systems (NIPS)}, 2014, pp. 568--576.

\bibitem{9.wu2015modeling}
Z.~Wu, X.~Wang, Y.-G. Jiang, H.~Ye, and X.~Xue, ``Modeling spatial-temporal
  clues in a hybrid deep learning framework for video classification,'' in
  \emph{ACM International Conference on Multimedia (ACM MM)}, 2015, pp.
  461--470.

\bibitem{87.zhao2017pooling}
S.~Zhao, Y.~Liu, Y.~Han, R.~Hong, Q.~Hu, and Q.~Tian, ``Pooling the
  convolutional layers in deep convnets for video action recognition,''
  \emph{IEEE Transactions on Circuits and Systems for Video Technology
  (TCSVT)}, 2017.

\bibitem{18.dalal2005histograms}
N.~Dalal and B.~Triggs, ``Histograms of oriented gradients for human
  detection,'' in \emph{IEEE Conference on Computer Vision and Pattern
  Recognition (CVPR)}, vol.~1, 2005, pp. 886--893.

\bibitem{115.dalal2006human}
N.~Dalal, B.~Triggs, and C.~Schmid, ``Human detection using oriented histograms
  of flow and appearance,'' in \emph{European conference on computer vision
  (ECCV)}, 2006, pp. 428--441.

\bibitem{lecun2015deep}
Y.~LeCun, Y.~Bengio, and G.~Hinton, ``Deep learning,'' \emph{Nature}, vol. 521,
  no. 7553, pp. 436--444, 2015.

\bibitem{bengio2013representation}
Y.~Bengio, A.~Courville, and P.~Vincent, ``Representation learning: A review
  and new perspectives,'' \emph{IEEE Transactions on Pattern Analysis and
  Machine Intelligence (TPAMI)}, vol.~35, no.~8, pp. 1798--1828, 2013.

\bibitem{57.wang2015action}
L.~Wang, Y.~Qiao, and X.~Tang, ``Action recognition with trajectory-pooled
  deep-convolutional descriptors,'' in \emph{IEEE Conference on Computer Vision
  and Pattern Recognition (CVPR)}, 2015, pp. 4305--4314.

\bibitem{xie2017hybrid}
G.-S. Xie, X.-Y. Zhang, S.~Yan, and C.-L. Liu, ``Hybrid cnn and
  dictionary-based models for scene recognition and domain adaptation,''
  \emph{IEEE Transactions on Circuits and Systems for Video Technology
  (TCSVT)}, vol.~27, no.~6, pp. 1263--1274, 2017.

\bibitem{23.yue2015beyond}
J.~Yue-Hei~Ng, M.~Hausknecht, S.~Vijayanarasimhan, O.~Vinyals, R.~Monga, and
  G.~Toderici, ``Beyond short snippets: Deep networks for video
  classification,'' in \emph{IEEE Conference on Computer Vision and Pattern
  Recognition (CVPR)}, 2015, pp. 4694--4702.

\bibitem{62.jaderberg2015spatial}
M.~Jaderberg, K.~Simonyan, A.~Zisserman \emph{et~al.}, ``Spatial transformer
  networks,'' in \emph{Advances in Neural Information Processing Systems
  (NIPS)}, 2015, pp. 2017--2025.

\bibitem{60.mnih2014recurrent}
V.~Mnih, N.~Heess, A.~Graves \emph{et~al.}, ``Recurrent models of visual
  attention,'' in \emph{Advances in Neural Information Processing Systems
  (NIPS)}, 2014, pp. 2204--2212.

\bibitem{50.zhao2008information}
Z.~Zhao and A.~M. Elgammal, ``Information theoretic key frame selection for
  action recognition,'' in \emph{British Machine Vision Conference (BMVC)},
  2008, pp. 1--10.

\bibitem{51.liu2013boosted}
L.~Liu, L.~Shao, and P.~Rockett, ``Boosted key-frame selection and correlated
  pyramidal motion-feature representation for human action recognition,''
  \emph{Pattern recognition (PR)}, vol.~46, no.~7, pp. 1810--1818, 2013.

\bibitem{2.wang2013action}
H.~Wang and C.~Schmid, ``Action recognition with improved trajectories,'' in
  \emph{IEEE International Conference on Computer Vision (ICCV)}, 2013, pp.
  3551--3558.

\bibitem{16.laptev2005space}
I.~Laptev, ``On space-time interest points,'' \emph{International Journal of
  Computer Vision (IJCV)}, vol.~64, no. 2-3, pp. 107--123, 2005.

\bibitem{90.xian2017evaluation}
Y.~Xian, X.~Rong, X.~Yang, and Y.~Tian, ``Evaluation of low-level features for
  real-world surveillance event detection,'' \emph{IEEE Transactions on
  Circuits and Systems for Video Technology (TCSVT)}, vol.~27, no.~3, pp.
  624--634, 2017.

\bibitem{wang2013mining}
L.~Wang, Y.~Qiao, and X.~Tang, ``Mining motion atoms and phrases for complex
  action recognition,'' in \emph{IEEE International Conference on Computer
  Vision (ICCV)}, 2013, pp. 2680--2687.

\bibitem{17.liu2011recognizing}
J.~Liu, B.~Kuipers, and S.~Savarese, ``Recognizing human actions by
  attributes,'' in \emph{IEEE Conference on Computer Vision and Pattern
  Recognition (CVPR)}, 2011, pp. 3337--3344.

\bibitem{wang2013motionlets}
L.~Wang, Y.~Qiao, and X.~Tang, ``Motionlets: Mid-level 3d parts for human
  motion recognition,'' in \emph{IEEE Conference on Computer Vision and Pattern
  Recognition (CVPR)}, 2013, pp. 2674--2681.

\bibitem{fradi2017crowd}
H.~Fradi, B.~Luvison, and Q.~C. Pham, ``Crowd behavior analysis using local
  mid-level visual descriptors,'' \emph{IEEE Transactions on Circuits and
  Systems for Video Technology (TCSVT)}, vol.~27, no.~3, pp. 589--602, 2017.

\bibitem{izadinia2012recognizing}
H.~Izadinia and M.~Shah, ``Recognizing complex events using large margin joint
  low-level event model,'' \emph{European Conference on Computer Vision
  (ECCV)}, pp. 430--444, 2012.

\bibitem{sun2013active}
C.~Sun and R.~Nevatia, ``Active: Activity concept transitions in video event
  classification,'' in \emph{IEEE International Conference on Computer Vision
  (ICCV)}, 2013, pp. 913--920.

\bibitem{89.zhang2017exploring}
Y.~Zhang, L.~Qin, R.~Ji, S.~Zhao, Q.~Huang, and J.~Luo, ``Exploring coherent
  motion patterns via structured trajectory learning for crowd mood modeling,''
  \emph{IEEE Transactions on Circuits and Systems for Video Technology
  (TCSVT)}, vol.~27, no.~3, pp. 635--648, 2017.

\bibitem{wu2016multi}
Z.~Wu, Y.-G. Jiang, X.~Wang, H.~Ye, and X.~Xue, ``Multi-stream multi-class
  fusion of deep networks for video classification,'' in \emph{ACM
  International Conference on Multimedia (ACM MM)}, 2016, pp. 791--800.

\bibitem{11.ji20133d}
S.~Ji, W.~Xu, M.~Yang, and K.~Yu, ``3d convolutional neural networks for human
  action recognition,'' \emph{IEEE Transactions on Pattern Analysis and Machine
  Intelligence (TPAMI)}, vol.~35, no.~1, pp. 221--231, 2013.

\bibitem{10.tran2014learning}
D.~Tran, L.~Bourdev, R.~Fergus, L.~Torresani, and M.~Paluri, ``Learning
  spatiotemporal features with 3d convolutional networks,'' in \emph{IEEE
  International Conference on Computer Vision (ICCV)}, 2015, pp. 4489--4497.

\bibitem{56.sun2015human}
L.~Sun, K.~Jia, D.-Y. Yeung, and B.~E. Shi, ``Human action recognition using
  factorized spatio-temporal convolutional networks,'' in \emph{IEEE
  International Conference on Computer Vision (ICCV)}, 2015, pp. 4597--4605.

\bibitem{wang2016temporal}
L.~Wang, Y.~Xiong, Z.~Wang, Y.~Qiao, D.~Lin, X.~Tang, and L.~Van~Gool,
  ``Temporal segment networks: Towards good practices for deep action
  recognition,'' in \emph{European Conference on Computer Vision (ECCV)}, 2016,
  pp. 20--36.

\bibitem{106.donahue2015long}
J.~Donahue, L.~Anne~Hendricks, S.~Guadarrama, M.~Rohrbach, S.~Venugopalan,
  K.~Saenko, and T.~Darrell, ``Long-term recurrent convolutional networks for
  visual recognition and description,'' in \emph{IEEE Conference on Computer
  Vision and Pattern Recognition (CVPR)}, 2015, pp. 2625--2634.

\bibitem{103.feichtenhofer2016spatiotemporal}
C.~Feichtenhofer, A.~Pinz, and R.~Wildes, ``Spatiotemporal residual networks
  for video action recognition,'' in \emph{Advances in Neural Information
  Processing Systems (NIPS)}, 2016, pp. 3468--3476.

\bibitem{yan2014modeling}
X.~Yan, H.~Chang, S.~Shan, and X.~Chen, ``Modeling video dynamics with deep
  dynencoder,'' in \emph{European Conference on Computer Vision (ECCV)}, 2014,
  pp. 215--230.

\bibitem{herath2017going}
S.~Herath, M.~Harandi, and F.~Porikli, ``Going deeper into action recognition:
  A survey,'' \emph{Image and Vision Computing (IVC)}, vol.~60, pp. 4--21,
  2017.

\bibitem{53.feichtenhofer2016convolutional}
C.~Feichtenhofer, A.~Pinz, and A.~Zisserman, ``Convolutional two-stream network
  fusion for video action recognition,'' in \emph{IEEE Conference on Computer
  Vision and Pattern Recognition (CVPR)}, 2016, pp. 1933--1941.

\bibitem{55.srivastava2015unsupervised}
N.~Srivastava, E.~Mansimov, and R.~Salakhutdinov, ``Unsupervised learning of
  video representations using lstms.'' in \emph{International Conference on
  Machine Learning (ICML)}, 2015, pp. 843--852.

\bibitem{goroshin2015unsupervised}
R.~Goroshin, J.~Bruna, J.~Tompson, D.~Eigen, and Y.~LeCun, ``Unsupervised
  learning of spatiotemporally coherent metrics,'' in \emph{IEEE International
  Conference on Computer Vision (ICCV)}, 2015, pp. 4086--4093.

\bibitem{wang2016actions}
X.~Wang, A.~Farhadi, and A.~Gupta, ``Actions\~{} transformations,'' in
  \emph{IEEE Conference on Computer Vision and Pattern Recognition (CVPR)},
  2016, pp. 2658--2667.

\bibitem{7.karpathy2014large}
A.~Karpathy, G.~Toderici, S.~Shetty, T.~Leung, R.~Sukthankar, and L.~Fei-Fei,
  ``Large-scale video classification with convolutional neural networks,'' in
  \emph{IEEE Conference on Computer Vision and Pattern Recognition (CVPR)},
  2014, pp. 1725--1732.

\bibitem{92.barrett2016action}
D.~P. Barrett and J.~M. Siskind, ``Action recognition by time series of
  retinotopic appearance and motion features,'' \emph{IEEE Transactions on
  Circuits and Systems for Video Technology (TCSVT)}, vol.~26, no.~12, pp.
  2250--2263, 2016.

\bibitem{lu2016hierarchical}
J.~Lu, J.~Yang, D.~Batra, and D.~Parikh, ``Hierarchical question-image
  co-attention for visual question answering,'' in \emph{Advances In Neural
  Information Processing Systems (NIPS)}, 2016, pp. 289--297.

\bibitem{su2017multi}
C.~Su, F.~Yang, S.~Zhang, Q.~Tian, L.~S. Davis, and W.~Gao, ``Multi-task
  learning with low rank attribute embedding for multi-camera person
  re-identification,'' \emph{IEEE Transactions on Pattern Analysis and Machine
  Intelligence (TPAMI)}, 2017.

\bibitem{tang2017collaborative}
Z.~Tang, L.~Li, D.~Wang, R.~Vipperla, Z.~Tang, L.~Li, D.~Wang, and R.~Vipperla,
  ``Collaborative joint training with multitask recurrent model for speech and
  speaker recognition,'' \emph{IEEE/ACM Transactions on Audio, Speech and
  Language Processing (TASLP)}, vol.~25, no.~3, pp. 493--504, 2017.

\bibitem{95.he2016deep}
K.~He, X.~Zhang, S.~Ren, and J.~Sun, ``Deep residual learning for image
  recognition,'' in \emph{IEEE Conference on Computer Vision and Pattern
  Recognition (CVPR)}, 2016, pp. 770--778.

\bibitem{li2016visual}
G.~Li and Y.~Yu, ``Visual saliency detection based on multiscale deep cnn
  features,'' \emph{IEEE Transactions on Image Processing (TIP)}, vol.~25,
  no.~11, pp. 5012--5024, 2016.

\bibitem{wang2015consistent}
W.~Wang, J.~Shen, and L.~Shao, ``Consistent video saliency using local gradient
  flow optimization and global refinement,'' \emph{IEEE Transactions on Image
  Processing (TIP)}, vol.~24, no.~11, pp. 4185--4196, 2015.

\bibitem{mauthner2015encoding}
T.~Mauthner, H.~Possegger, G.~Waltner, and H.~Bischof, ``Encoding based
  saliency detection for videos and images,'' in \emph{IEEE Conference on
  Computer Vision and Pattern Recognition (CVPR)}, 2015, pp. 2494--2502.

\bibitem{52.zhou2016learning}
B.~Zhou, A.~Khosla, A.~Lapedriza, A.~Oliva, and A.~Torralba, ``Learning deep
  features for discriminative localization,'' in \emph{IEEE Conference on
  Computer Vision and Pattern Recognition (CVPR)}, 2016, pp. 2921--2929.

\bibitem{yang2016discovering}
J.~Yang, G.~Zhao, J.~Yuan, X.~Shen, Z.~Lin, B.~Price, and J.~Brandt,
  ``Discovering primary objects in videos by saliency fusion and iterative
  appearance estimation,'' \emph{IEEE Transactions on Circuits and Systems for
  Video Technology (TCSVT)}, vol.~26, no.~6, pp. 1070--1083, 2016.

\bibitem{liu2017global}
J.~Liu, G.~Wang, P.~Hu, L.-Y. Duan, and A.~C. Kot, ``Global context-aware
  attention lstm networks for 3d action recognition,'' in \emph{IEEE Conference
  on Computer Vision and Pattern Recognition (CVPR)}, 2017, pp. 1647--1656.

\bibitem{116.karmarkar1984new}
N.~Karmarkar, ``A new polynomial-time algorithm for linear programming,'' in
  \emph{ACM Symposium on Theory of Computing (STOC)}, 1984, pp. 302--311.

\bibitem{93.kuehne2011hmdb}
H.~Kuehne, H.~Jhuang, E.~Garrote, T.~Poggio, and T.~Serre, ``Hmdb: a large
  video database for human motion recognition,'' in \emph{IEEE International
  Conference on Computer Vision (ICCV)}, 2011, pp. 2556--2563.

\bibitem{94.reddy2013recognizing}
K.~K. Reddy and M.~Shah, ``Recognizing 50 human action categories of web
  videos,'' \emph{Machine Vision and Applications}, vol.~24, no.~5, pp.
  971--981, 2013.

\bibitem{36.soomro2012ucf101}
K.~Soomro, A.~R. Zamir, and M.~Shah, ``Ucf101: A dataset of 101 human actions
  classes from videos in the wild,'' \emph{Technical Report CRCV-TR-12-01},
  2012.

\bibitem{THUMOS14}
Y.-G. Jiang, J.~Liu, A.~Roshan~Zamir, G.~Toderici, I.~Laptev, M.~Shah, and
  R.~Sukthankar, ``{THUMOS} challenge: Action recognition with a large number
  of classes,'' \url{http://crcv.ucf.edu/THUMOS14/}, 2014.

\bibitem{wang2017untrimmednets}
L.~Wang, Y.~Xiong, D.~Lin, and L.~Van~Gool, ``Untrimmednets for weakly
  supervised action recognition and detection,'' \emph{arXiv preprint
  arXiv:1703.03329}, 2017.

\bibitem{jain201515}
M.~Jain, J.~C. van Gemert, and C.~G. Snoek, ``What do 15,000 object categories
  tell us about classifying and localizing actions?'' in \emph{IEEE Conference
  on Computer Vision and Pattern Recognition (CVPR)}, 2015, pp. 46--55.

\bibitem{zach2007duality}
C.~Zach, T.~Pock, and H.~Bischof, ``A duality based approach for realtime tv-l
  1 optical flow,'' \emph{Pattern Recognition (PR)}, pp. 214--223, 2007.

\bibitem{3.krizhevsky2012imagenet}
A.~Krizhevsky, I.~Sutskever, and G.~E. Hinton, ``Imagenet classification with
  deep convolutional neural networks,'' in \emph{Advances in Neural Information
  Processing Systems (NIPS)}, 2012, pp. 1097--1105.

\bibitem{simonyan15verydeep}
K.~Simonyan and A.~Zisserman, ``Very deep convolutional networks for
  large-scale image recognition,'' in \emph{International Conference on
  Learning Representations (ICLR)}, 2015, pp. 1--14.

\bibitem{szegedy2015going}
C.~Szegedy, W.~Liu, Y.~Jia, P.~Sermanet, S.~Reed, D.~Anguelov, D.~Erhan,
  V.~Vanhoucke, and A.~Rabinovich, ``Going deeper with convolutions,'' in
  \emph{IEEE Conference on Computer Vision and Pattern Recognition (CVPR)},
  2015, pp. 1--9.

\bibitem{112.cai2014multi}
Z.~Cai, L.~Wang, X.~Peng, and Y.~Qiao, ``Multi-view super vector for action
  recognition,'' in \emph{IEEE Conference on Computer Vision and Pattern
  Recognition (CVPR)}, 2014, pp. 596--603.

\bibitem{114.wang2013dense}
H.~Wang, A.~Kl{\"a}ser, C.~Schmid, and C.-L. Liu, ``Dense trajectories and
  motion boundary descriptors for action recognition,'' \emph{International
  Journal of Computer Vision (IJCV)}, vol. 103, no.~1, pp. 60--79, 2013.

\bibitem{98.lan2015beyond}
Z.~Lan, M.~Lin, X.~Li, A.~G. Hauptmann, and B.~Raj, ``Beyond gaussian pyramid:
  Multi-skip feature stacking for action recognition,'' in \emph{IEEE
  Conference on Computer Vision and Pattern Recognition (CVPR)}, 2015, pp.
  204--212.

\bibitem{108.wang2016deep}
J.~Wang, W.~Wang, R.~Wang, W.~Gao \emph{et~al.}, ``Deep alternative neural
  network: Exploring contexts as early as possible for action recognition,'' in
  \emph{Advances in Neural Information Processing Systems (NIPS)}, 2016, pp.
  811--819.

\bibitem{101.chen2010long}
M.-y. Chen, ``Long term activity analysis in surveillance video archives,''
  Ph.D. dissertation, Carnegie Mellon University, 2010.

\bibitem{105.zhu2016key}
W.~Zhu, J.~Hu, G.~Sun, X.~Cao, and Y.~Qiao, ``A key volume mining deep
  framework for action recognition,'' in \emph{IEEE Conference on Computer
  Vision and Pattern Recognition (CVPR)}, 2016, pp. 1991--1999.

\bibitem{yang2017deep}
Y.~Yang, D.-C. Zhan, Y.~Fan, Y.~Jiang, and Z.-H. Zhou, ``Deep learning for
  fixed model reuse,'' in \emph{AAAI Conference on Artificial Intelligence
  (AAAI)}, 2017, pp. 2831--2837.

\bibitem{109.bilen2016dynamic}
H.~Bilen, B.~Fernando, E.~Gavves, A.~Vedaldi, and S.~Gould, ``Dynamic image
  networks for action recognition,'' in \emph{IEEE Conference on Computer
  Vision and Pattern Recognition (CVPR)}, 2016, pp. 3034--3042.

\bibitem{100.peng2016bag}
X.~Peng, L.~Wang, X.~Wang, and Y.~Qiao, ``Bag of visual words and fusion
  methods for action recognition: Comprehensive study and good practice,''
  \emph{Computer Vision and Image Understanding (CVIU)}, vol. 150, pp.
  109--125, 2016.

\bibitem{104.li2016vlad3}
Y.~Li, W.~Li, V.~Mahadevan, and N.~Vasconcelos, ``Vlad3: Encoding dynamics of
  deep features for action recognition,'' in \emph{IEEE Conference on Computer
  Vision and Pattern Recognition (CVPR)}, 2016, pp. 1951--1960.

\bibitem{110.fernando2015modeling}
B.~Fernando, E.~Gavves, J.~M. Oramas, A.~Ghodrati, and T.~Tuytelaars,
  ``Modeling video evolution for action recognition,'' in \emph{IEEE Conference
  on Computer Vision and Pattern Recognition (CVPR)}, 2015, pp. 5378--5387.

\bibitem{99.wang2016robust}
H.~Wang, D.~Oneata, J.~Verbeek, and C.~Schmid, ``A robust and efficient video
  representation for action recognition,'' \emph{International Journal of
  Computer Vision (IJCV)}, vol. 119, no.~3, pp. 219--238, 2016.

\bibitem{varol2015efficient}
G.~Varol and A.~A. Salah, ``Efficient large-scale action recognition in videos
  using extreme learning machines,'' \emph{Expert Systems with Applications
  (ESWA)}, vol.~42, no.~21, pp. 8274--8282, 2015.

\bibitem{96.oneata2013action}
D.~Oneata, J.~Verbeek, and C.~Schmid, ``Action and event recognition with
  fisher vectors on a compact feature set,'' in \emph{IEEE International
  Conference on Computer Vision (ICCV)}, 2013, pp. 1817--1824.

\bibitem{zhang2016real}
B.~Zhang, L.~Wang, Z.~Wang, Y.~Qiao, and H.~Wang, ``Real-time action
  recognition with enhanced motion vector cnns,'' in \emph{IEEE Conference on
  Computer Vision and Pattern Recognition (CVPR)}, 2016, pp. 2718--2726.

\bibitem{113.ciptadi2014movement}
A.~Ciptadi, M.~S. Goodwin, and J.~M. Rehg, ``Movement pattern histogram for
  action recognition and retrieval,'' in \emph{European Conference on Computer
  Vision (ECCV)}, 2014, pp. 695--710.

\bibitem{111.wu2014towards}
J.~Wu, Y.~Zhang, and W.~Lin, ``Towards good practices for action video
  encoding,'' in \emph{IEEE Conference on Computer Vision and Pattern
  Recognition (CVPR)}, 2014, pp. 2577--2584.

\bibitem{97.narayan2014cause}
S.~Narayan and K.~R. Ramakrishnan, ``A cause and effect analysis of motion
  trajectories for modeling actions,'' in \emph{IEEE Conference on Computer
  Vision and Pattern Recognition (CVPR)}, 2014, pp. 2633--2640.

\bibitem{117.kloft2011lp}
M.~Kloft, U.~Brefeld, S.~Sonnenburg, and A.~Zien, ``Lp-norm multiple kernel
  learning,'' \emph{Journal of Machine Learning Research (JMLR)}, vol.~12, no.
  Mar, pp. 953--997, 2011.

\bibitem{Vedaldi2009multiple}
A.~Vedaldi, V.~Gulshan, M.~Varma, and A.~Zisserman, ``Multiple kernels for
  object detection,'' in \emph{IEEE International Conference on Computer Vision
  (ICCV)}, 2009, pp. 606--613.

\end{thebibliography}

\begin{IEEEbiography}[{\includegraphics[width=1in,height=1.25in,clip,keepaspectratio]{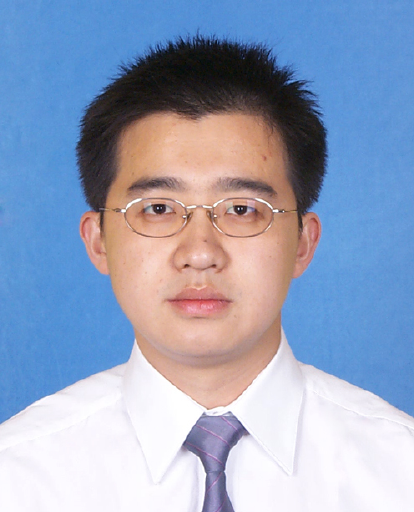}}]
{Yuxin Peng} is the professor of Institute of Computer Science and Technology (ICST), Peking University, and the chief scientist of 863 Program (National Hi-Tech Research and Development Program of China). He received the Ph.D. degree in computer application from Peking University in Jul. 2003. After that, he worked as an assistant professor in ICST, Peking University. He was promoted to associate professor and professor in Peking University in Aug. 2005 and Aug. 2010 respectively. In 2006, he was authorized by the ``Program for New Century Excellent Talents in University (NCET)'' and the ``Program for New Star in Science and Technology of Beijing''. He has published more than 100 papers in refereed international journals and conference proceedings, including IJCV, TIP, TMM, TCSVT, PR, ACM MM, ICCV, CVPR, IJCAI, AAAI, etc. He led his team to participate in TRECVID (TREC Video Retrieval Evaluation) many times. In TRECVID 2009, his team won four first places on 4 sub-tasks of the High-Level Feature Extraction (HLFE) task and Search task. In TRECVID 2012, his team gained four first places on all 4 sub-tasks of the Instance Search (INS) task and Known-Item Search (KIS) task. In TRECVID 2014, his team gained the first place in the Interactive Instance Search task. His team also gained both two first places in the INS task of TRECVID 2015, 2016 and 2017. Besides, he received the first prize of Beijing Science and Technology Award for Technological Invention in 2016 (ranking first). He has applied 35 patents, and obtained 16 of them. His current research interests mainly include cross-media analysis and reasoning, image and video analysis and retrieval, and computer vision.
\end{IEEEbiography}
\begin{IEEEbiography}[{\includegraphics[width=1in,height=1.25in,clip,keepaspectratio]{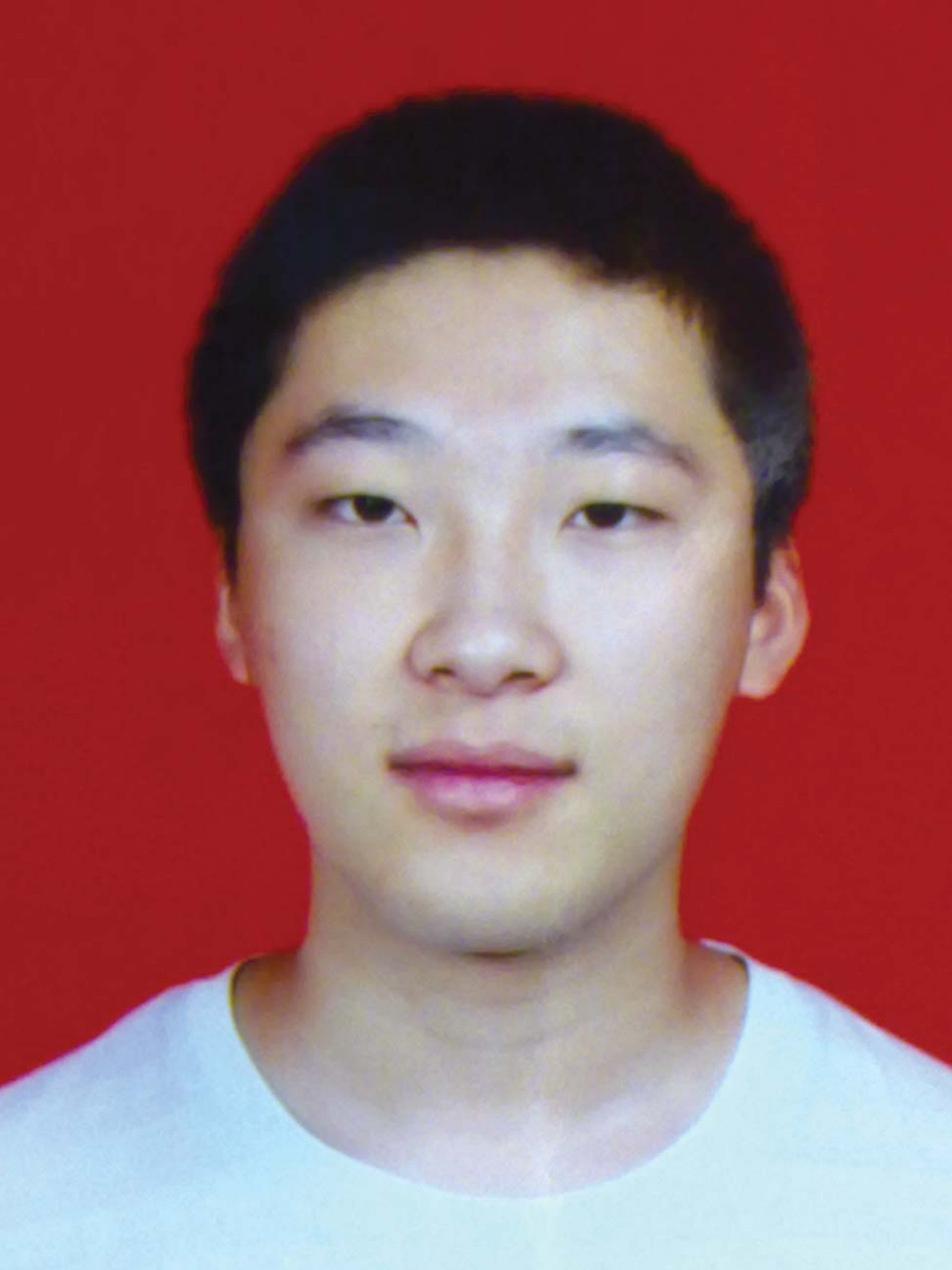}}]
{Yunzhen Zhao} received the B.S. degree in mathematical science from Peking University, in July 2014. He is currently pursuing the M.S. degree in the Institute of Computer Science and Technology (ICST), Peking University. His current research interests include cross-media retrieval and machine learning.
\end{IEEEbiography}
\begin{IEEEbiography}[{\includegraphics[width=1in,height=1.25in,clip,keepaspectratio]{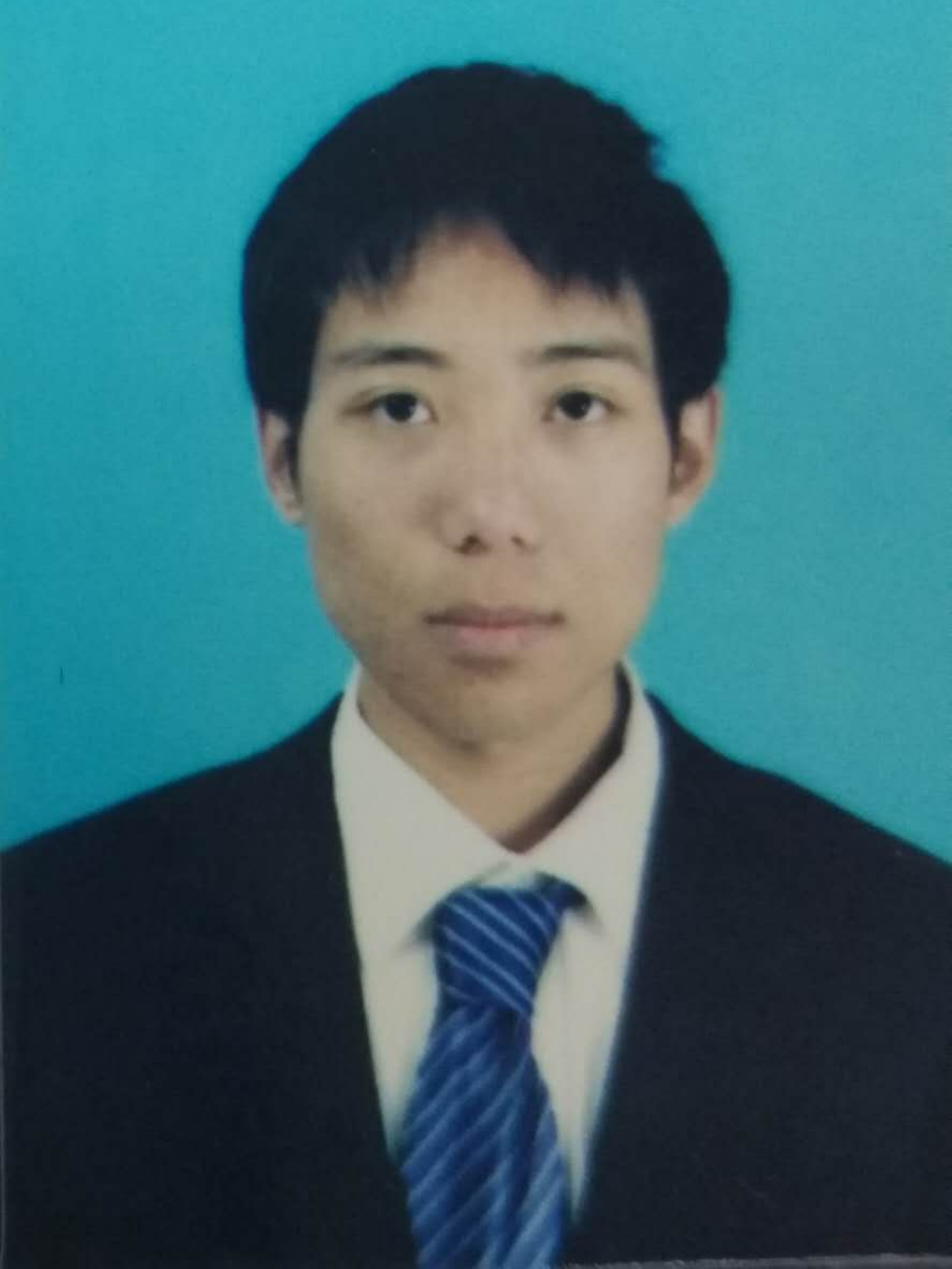}}]
{Junchao Zhang} reveived the B.S. degree in computer science from Nankai University, Tianjin, China, in 2015. He is currently working toward the Ph.D. degree in the Institute of Computer Science and Technology, Peking University, Beijing, China.
His research interests include video analysis and recognition, and machine learning.
\end{IEEEbiography}
\end{document}